# Towards Responsible and Trustworthy Educational Data Mining: Comparing Symbolic, Sub-Symbolic, and Neural-Symbolic AI Methods

Danial Hooshyar[1,2*], Eve Kikas[3], Yeongwook Yang[4], Gustav Šír[5], Raija Hämäläinen[2], Tommi Kärkkäinen[6], Roger Azevedo[7]

[1]School of digital Technologies, Tallinn University, Tallinn, Estonia
[2]Department of Education, University of Jyväskylä, Jyväskylä, Finland
[3]School of Natural Sciences and Health, Tallinn University, Tallinn, Estonia
[4]Department of Computer Science and Engineering, Gangneung-Wonju National University Wonju, Republic of Korea
[5]Department of Computer Science, Czech Technical University, Prague, the Czech Republic
[6]Faculty of Information Technology, University of Jyväskylä, Jyväskylä, Finland
[7]School of Modeling Simulation and Training, University of Central Florida, Orlando, USA
* Corresponding author: danial.hooshyar@tlu.ee

**Abstract**

Given the demand for responsible and trustworthy AI for education, this study evaluates symbolic, sub-symbolic, and neural-symbolic AI (NSAI) in terms of generalizability and interpretability. Our extensive experiments on balanced and imbalanced self-regulated learning datasets of Estonian primary school students predicting 7th-grade mathematics national test performance showed that symbolic and sub-symbolic methods performed well on balanced data but struggled to identify low performers in imbalanced datasets. Interestingly, symbolic and sub-symbolic methods emphasized different factors in their decision-making: symbolic approaches primarily relied on cognitive and motivational factors, while sub-symbolic methods focused more on cognitive aspects, learnt knowledge, and the demographic variable of gender—yet both largely overlooked metacognitive factors. The NSAI method, on the other hand, showed advantages by: (i) being more generalizable across both classes—even in imbalanced datasets—as its symbolic knowledge component compensated for the underrepresented class; and (ii) relying on a more integrated set of factors in its decision-making, including motivation, (meta)cognition, and learnt knowledge, thus offering a comprehensive and theoretically grounded interpretability framework. These contrasting findings highlight the need for a holistic comparison of AI methods before drawing conclusions based solely on predictive performance. They also underscore the potential of hybrid, human-centred NSAI methods to address the limitations of other AI families and move us closer to responsible AI for education. Specifically, by enabling stakeholders to contribute to AI design, NSAI aligns learned patterns with theoretical constructs, incorporates factors like motivation and metacognition, and strengthens the trustworthiness and responsibility of educational data mining.

**Keywords** Artificial intelligence in education . Responsible and trustworthy AI . Explainable AI . Educational data mining . Learning analytics

## 1. Introduction

Educational data mining (EDM) is an interdisciplinary field that applies computational techniques to analyse educational data and uncover meaningful patterns that can enhance teaching and learning (Romero et al., 2010). By leveraging artificial intelligence (AI)-driven methods, e.g., supervised and unsupervised machine learning, EDM helps in grouping or predicting student performance (Hooshyar et al., 2020, 2023; Vellido et al., 2006; Yağcı, 2022), personalizing learning pathways (Lin et al., 2013; McCalla, 2004), detecting students at risk of failure (Hooshyar et al., 2022; Sarra et al., 2019), and improving adaptive learning environments (Koedinger et al., 2015; Romero & Ventura, 2007). AI models in EDM can extract valuable insights from student interactions, assessments, and behavioural data, providing educators with data-driven strategies for better decision-making (Peña-Ayala, 2014). Traditionally, predictive modelling in EDM has relied on symbolic and sub-symbolic approaches, each with distinct advantages and limitations (Holmes et al., 2022).

Symbolic methods rely on structured, human-readable representations that explicitly encode relationships in the data (Ilkou & Koutraki, 2020). They offer transparent, step-by-step reasoning, making them easier to control, explain, and troubleshoot—an essential feature for educational settings where trust and interpretability matter (Conati et al., 2018). Additionally, symbolic methods, particularly rule-based systems, provide modularity, allowing individual rule modifications without disrupting the overall system, making them well-suited for structured educational problems (Hatzilygeroudis & Prentzas, 2004). However, symbolic methods have also notable limitations. They often struggle to capture complex relationships and lack mechanisms for handling uncertainty and noise in real-world educational data, limiting their generalization capabilities (Garcez et al., 2022; Lenat et al., 1985). Moreover, the reliance on symbolic methods introduces the knowledge acquisition bottleneck—a significant challenge due to the high cost of encoding real-world educational problems into symbolic representations (Cullen & Bryman, 1988). This makes symbolic AI less adaptable to evolving learning environments, as updating the symbolic domain language requires human intervention. An example of this AI family includes Lim et al. (2023)'s work, which applied a rule-based technique to extract patterns from trace data using expert-defined "IF [condition] THEN [act]" rules to detect self-regulated learning processes and trigger real-time personalized scaffolds (Azevedo et al., 2024). By first labelling raw trace data as learning actions and mapping sequences of actions to self-regulated learning activities, the system identified patterns in student behaviour to determine when scaffolds were needed and which ones to trigger, depending on a number of conditions (Azevedo et al., 2022).

On the other hand, sub-symbolic methods, such as support vector machine and neural networks, excel at predictive performance by learning complex statistical relationships from data. Unlike symbolic methods, which allow human supervision and explicit rule-based reasoning, sub-symbolic approaches establish direct mappings between input and output variables, often through mathematical functions or probabilistic models (Holmes et al., 2022). Bayesian networks offer a hybrid approach, combining expert knowledge with data-driven learning, making them particularly effective for latent variable modelling, which is essential for tracking students' evolving processes (Hooshyar & Druzdzel, 2024). However, Bayesian networks are computationally expensive and/or require expert-defined structures (Daly et al., 2011). Neural networks, despite their adaptability, ability to model non-linear (sequential) patterns, and often superior performance, suffer from poor interpretability, making it difficult to explain AI-driven educational insights to educators and learners (Besold et al., 2021; Garcez et al., 2019). Yet, one can provide model-specific information on feature importance, similar to random forests (Linja et al., 2023). Additionally, sub-symbolic methods are highly dependent on training data, meaning they may fail to generalize when exposed to lack of data or underrepresented student behaviours (Tato & Nkambou, 2022; Torralba & Efros, 2011). An example of sub-symbolic methods in EDM includes Altaf et al. (2019)'s use of neural networks to predict student performance from learning management system data. Trained on Moodle logs from 900 students, neural networks outperformed six classifiers in accuracy and matched the best in recall. The study found prior grades to be the strongest predictor, highlighting the effectiveness of neural networks as a sub-symbolic method in EDM. For a systematic review of neural networks in EDM, see Okewu et al. (2021).

While sub-symbolic methods, particularly deep neural networks, often achieve superior predictive accuracy, they present significant challenges in educational applications. Unlike symbolic models, they lack explicit domain knowledge encoding, making it harder to control their decision-making process in a way that aligns AI insights with educational frameworks. These models rely on numerical data, requiring *educational knowledge* (e.g., educational or psychological constructs like SRL, causal relationships, and practitioners' knowledge) to be converted into numerical representations, which can reduce semantic richness and structural coherence (Dash et al., 2022). This limitation affects their ability to fully capture the complexity of learning processes and student behaviour in a way that aligns with educational standards and guidelines (Garcez et al., 2019; Hooshyar et al., 2024). While they excel at pattern detection, they struggle to understand and integrate cause-and-effect relationships in learning contexts (Kitto et al., 2023). For example, when modelling variables like gender and self-efficacy, they may misinterpret causality, incorrectly treating self-efficacy as a predictor of gender. Integrating symbolic knowledge can mitigate these challenges, particularly when representative training data is limited.

Moreover, deep neural network models have a tendency to learn and perpetuate biases by acquiring misleading correlations during training, leading to dependence on irrelevant features and reduced generalizability (Y.-Y. Yang et al., 2022). In many cases, educational datasets may be limited or fail to fully reflect diverse learning scenarios and student behaviours, causing deep neural networks to struggle to generalize accurately when faced with limited or biased information (R. S. Baker & Hawn, 2022; Tato & Nkambou, 2022; Torralba & Efros, 2011). Misleading correlations in AI models can reinforce inequalities, misguide interventions, and fail to adapt to individual learners, undermining equitable education (Vincent-Lancrin & Van der Vlies, 2020). For example, recent research showed that deep neural networks, when trained only on educational training data, tend to rely on spurious correlations, neglecting crucial causal features in decision-making, especially when they depend on data size or encounter unbalanced datasets (Hooshyar et al., 2024; Tato & Nkambou, 2022). Another major challenge of these methods is their *lack of interpretability* (Okewu et al., 2021). Interpretability refers to a model's ability to present its decision-making process in a way that humans can understand. In education, this is essential for multiple reasons (Conati et al., 2018; Fiok et al., 2022; Hooshyar & Yang, 2021). Interpretability fosters trust in deep neural networks, enabling meaningful feedback and critical thinking, and the interpretation should align with theoretically-based constructs and assumptions. It also helps detect biases, ensuring fairness and equal opportunities in education. Without it, AI's educational potential remains constrained. Recent research has explored explainer algorithms like SHAP (Lundberg & Lee, 2017) and LIME (Ribeiro et al., 2016) to improve interpretability in sub-symbolic methods for education (Hooshyar et al., 2022; Saarela et al., 2021). However, these approaches come with limitations. For instance, they may rely on a feature for predictions without explicitly including it in explanations, resulting in misleading interpretations (Lakkaraju & Bastani, 2020; Slack et al., 2020). Moreover, some of these explainable AI methods (including LIME) have moved away from fidelity as a measure of explanation quality, leading to explanations that may not accurately reflect the underlying model (Hooshyar & Yang, 2024; Rudin, 2019; White & d'Avila Garcez, 2020). For example, research conducted by Hooshyar and Yang (2024) have thoroughly examined the inner mechanisms of deep neural networks in education by extracting knowledge from trained models and comparing it with explanations from SHAP and LIME, exposing major inconsistencies in reliability and fidelity. Importantly, these explainable AI methods fail to capture the actual decision-making mechanisms and instead rely primarily on feature importance and model proxies to explain predictions.

### 1.1. Neural-symbolic computing: A step towards responsible and trustworthy educational data mining

To overcome these limitations, neural-symbolic AI (NSAI)[1], a hybrid method often referred to as the "third wave of AI", aims to integrate the strengths of both symbolic and sub-symbolic methods, combining the structured, interpretable knowledge representation of symbolic models with the powerful pattern-learning capabilities of neural networks (d'Avila Garcez & Lamb, 2020). By embedding symbolic knowledge into neural learning architectures (Sourek et al., 2018), NSAI allows models to both learn from data and incorporate domain-specific educational theories, ensuring that the discovered patterns align more closely with the pedagogical principles (Tato & Nkambou, 2022). This hybrid approach, unlike purely data-driven approaches, not only can potentially improve model's generalizability but also addresses key challenges in interpretability, trustworthiness, and bias mitigation—critical factors in high-stakes educational applications (Garcez et al., 2019; Hooshyar, 2024; Venugopal et al., 2021). This capability is particularly crucial given the increasing regulatory landscape surrounding AI in education. The recent designation of education as a high-risk area under the EU AI Act (European Union, 2024) necessitates *responsible AI systems* that balance predictive accuracy, transparency, and fairness. As Rudin (2019) argues, responsible AI requires that black-box models should not be used for high-stakes decisions if an equally performant interpretable model is available, as relying on opaque systems in such contexts raises ethical (Durán & Jongsma, 2021; Lo Piano, 2020) and legal (European Union, 2024) concerns. Given that AI in education falls under high-risk AI applications, interpretability is no longer just a technical goal but an ethical necessity, reinforcing the importance of NSAI in responsible educational data mining. NSAI offers a direct way to explain decision-making processes without relying on post-hoc explanations, reducing the need for black-box justifications and ensuring that AI-driven educational interventions remain accountable and trustworthy (Hooshyar et al., 2024; Hooshyar & Yang, 2024).

---

[1] The research community has traditionally used the term "neural-symbolic" to describe the integration of neural and symbolic AI, while "neurosymbolic" has also been used interchangeably in academic literature and the media.

A prominent example of this hybrid approach is the recent study conducted by Hooshyar et al. (2024), comparing deep neural networks with a knowledge-based deep neural network for predicting learner performance. NSAI, which integrates raw data and symbolic knowledge, demonstrated better generalizability and captured causal relationships more effectively than traditional models, even when data augmentation techniques like SMOTE (Chawla et al., 2002) and autoencoders (Baldi, 2012; Kärkkäinen & Hänninen, 2023) were used. The study highlighted NSAI's ability to extract meaningful rules, enhancing interpretability, structured reasoning, and hypothesis testing, contributing to responsible and trustworthy educational data mining.

The concept of responsible AI has been articulated through various frameworks, each emphasizing key principles such as fairness, privacy, accountability, and transparency. For instance, Maree et al. (2020) highlight fairness, privacy, accountability, transparency, and soundness, while Arrieta et al. (2020) expand this view to include ethics, security, and safety. Other perspectives, such as those of Eitel-Porter (2020) and Werder et al. (2022), stress explainability as a crucial component. A broader perspective is offered by Jakesch et al. (2022), incorporating sustainability, inclusiveness, social good, human autonomy, and solidarity into the responsible AI discourse. Building on these foundations, Goellner et al. (2024) conducted a systematic review of 254 studies and proposed a comprehensive yet practical definition: "*Responsible AI is **human-centred** and ensures users' **trust** through **ethical** ways of decision making. The decision-making must be fair, accountable, not biased, with good intentions, non-discriminating, and consistent with societal laws and norms. Responsible AI ensures, that automated decisions are **explainable** to users while always preserving users' **privacy** through a **secure** implementation.*"

NSAI aligns closely with responsible AI principles by integrating human-centred design, trust, ethics, explainability, privacy, and security into educational data mining. By embedding symbolic knowledge, it ensures AI-driven decisions are not purely data-driven but grounded in educational theories and ethical guidelines, fostering trust and human-centred AI. It also allows direct stakeholder involvement, enabling educators and practitioners to contribute their expertise to model training. Additionally, by incorporating structured knowledge, NSAI can compensate for data inconsistencies and underrepresented groups, mitigating bias and enhancing fairness. Its interpretable structure strengthens explainability, allowing educators, students, and policymakers to understand and validate model outputs, ensuring AI decisions align with pedagogical reasoning. Unlike black-box deep learning models, NSAI does not rely solely on post-hoc explanations, reducing the risk of misleading or opaque justifications. Besides overseeing model training to ensure it is conducted in a trustworthy and ethical manner, practitioners and learning science researchers can also gain insights into the actual logic behind AI decisions. This multi-level explainability, from fine to coarse-grained explanations, not only augments domain knowledge but also supports hypothesis testing, helping researchers refine educational theories using AI-driven insights. Furthermore, NSAI reinforces privacy and security by leveraging symbolic reasoning to enforce access control, auditability, and compliance with educational policies. By balancing predictive power with interpretability, fairness, and transparency, NSAI provides a responsible AI framework that supports equitable and trustworthy educational applications.

### 1.2. Self-regulated learning in predicting Math performance

By analysing self-regulated learning (SRL) indicators, this study demonstrates how symbolic, sub-symbolic, and neural-symbolic AI perform in balancing generalizability and interpretability, providing insights into their effectiveness in predicting mathematical performance. To understand how SRL is conceptualized and operationalized within the context of responsible and trustworthy AI in education, we first discuss its core principles. Various SRL models incorporate regulation of cognition, motivation, affect/emotions, behaviours, and context, although emphasizing these aspects somewhat differently (Greene et al., 2024; Panadero, 2017; Winne & Azevedo, 2022).

To support SRL competencies in Estonian schools (also called "learning to learn" competencies by the Estonian government), the model described by Kikas et al. (2023) is utilized. Recently, it was also elaborated for the digital environment (E. Kikas et al., 2024). This model is heavily based on the metacognitive and affective model of self-regulated learning, MARSL, which differentiates between Task and Person and emphasizes that learning occurs at the "Task × Person" level (Efklides, 2011; Efklides & Schwartz, 2024). The Person level encompasses the learner's current (meta)cognitive knowledge and skills, motivational beliefs, metamotivational knowledge and skills, as well as (meta)emotional characteristics that affect the entire learning process and outcomes. While Task × Person level activities can be assessed in a digital environment in real time during learning, Person-level indicators are general trait-like tendencies that influence learning across various tasks and contexts. The learner's Person-level characteristics affect the SRL process by interacting with the characteristics of a specific task at the Task × Person level (Efklides, 2011; Efklides & Schwartz, 2024).

Math skills develop cumulatively, meaning that students must first acquire basic knowledge and operations that serve as the foundation for more advanced skills (Korpipää et al., 2017). Thus, prior math skills play the most crucial role in later mathematical development (Watts et al., 2014). Moreover, two types of mathematical knowledge have been distinguished (Rittle-Johnson, 2017). Factual/procedural knowledge refers to knowing specific procedures, such as the steps or actions needed to reach a goal, while conceptual knowledge involves understanding abstract concepts and general principles. Studies have confirmed that stronger SRL competencies are associated with higher math performance (Dent & Koenka, 2016). However, the potential differences in how SRL competencies influence factual/procedural (later referred to as factual) versus conceptual knowledge have not yet been examined.

### 1.3. Aim of the study

SRL competencies provide valuable insights into students' learning, making their effective modelling crucial for personalized interventions and data-driven decision-making (Zhidkikh et al., 2023). However, existing AI-driven methods in educational data mining have primarily relied on either symbolic or sub-symbolic approaches, both of which have limitations in capturing the complexity of SRL.

Given the growing need for responsible and trustworthy AI in education, this study evaluates symbolic, sub-symbolic, and neural-symbolic AI in educational data mining from both quantitative (generalizability) and qualitative (interpretability) perspectives. Specifically, it uses SRL indicators and learnt knowledge from Grade 3 to predict national math test outcomes in Grade 7, demonstrating how neural-symbolic AI balances generalizability and interpretability while aligning with educational standards. SRL Person-level indicators were assessed using a web-based tool commonly implemented in Estonian schools (Kikas, 2018). This tool includes not only questionnaires but also specific tasks with follow-up questions to capture a more comprehensive profile of students' SRL competencies (Kikas & Jõgi, 2016; Rovers et al., 2019). The national math test, used as the prediction target, evaluates both factual/procedural and conceptual knowledge, ensuring a robust assessment of students' mathematical competencies.

## 2. Related Work

Educational Data Mining (EDM) focuses on extracting meaningful insights from educational data to enhance learning environments and improve educational systems (R. Baker, 2010; Romero & Ventura, 2013). A wide range of AI and machine learning techniques have been employed in this domain, including unsupervised methods such as clustering and association rule mining, as well as supervised approaches like classification and regression (R. Baker, 2010; Peña-Ayala, 2014; Romero & Ventura, 2007). Additionally, other methods including process mining have been explored to analyse learning behaviours and optimize instructional strategies (Bogarín et al., 2018).

In the field of predictive analytics, researchers have utilized various methods to predict student performance, including classifying learners (Albreiki et al., 2021), predicting grades (Meier et al., 2015; Saarela et al., 2016), detecting early dropout risks (Hooshyar et al., 2022; Nagy & Molontay, 2023; Zhidkikh et al., 2024), and personalizing learning pathways (Lin et al., 2013). The majority of these methods fall under two broad AI categories: symbolic AI and sub-symbolic AI. For instance, Yang et al. (2020) applied both AI families, utilizing unsupervised and supervised learning methods such as clustering, classification, and regression, to predict student outcomes in a Moodle-based course at the University of Tartu in Estonia. They first constructed feature vectors representing students' procrastination behaviours, incorporating factors such as active, inactive, and spare time for homework, along with homework grades. Using clustering techniques, they categorized students based on their procrastination tendencies. Subsequently, they employed classification and regression models to predict students' course achievement based on these behavioural patterns. Their findings highlight that some sub-symbolic methods, like support vector machines, outperformed symbolic methods, like decision trees. Similarly, multiple studies have reported that sub-symbolic models outperform symbolic AI methods in various educational tasks (Hellas et al., 2018; Hooshyar et al., 2019; Ibrahim & Rusli, 2007; Y. Yang et al., 2020). Neural networks are among the best-performing sub-symbolic methods, demonstrating great capability and high accuracy in predictive modelling tasks in education (Hernández-Blanco et al., 2019; Okewu et al., 2021). However, they face major challenges, including a tendency to learn and perpetuate biases by acquiring misleading correlations during training (Hooshyar et al., 2024), and a lack of interpretability (Durán & Jongsma, 2021; Hooshyar & Yang, 2024). While there have been some attempts to improve transparency through feature sensitivity-based explanation methods for distance-based supervised models (Linja et al., 2023), this remains a core concern in responsible AI, particularly in educational decision-making (Rudin, 2019).

To address such challenges, researchers have begun exploring neural-symbolic computing. For instance, Shakya et al. (2021) presented an approach that integrates symbolic educational knowledge with deep learning for student strategy prediction. Their hybrid model combined Markov Logic (Pedro & Lowd, 2009) to encode domain relationships with LSTMs (Hochreiter, 1997), improving training efficiency and reducing overfitting through importance sampling. Empirical evaluation on KDD EDM challenge datasets demonstrated the scalability of this approach. Telesko et al. (2020) proposed an NSAI method that leveraged Kohonen Feature Maps to cluster students based on their object-orientation skills and applied a post-processing component that extracted propositional rules from the trained network. This method enhanced interpretability by identifying meaningful student clusters and deriving fuzzy rules that aid in optimizing admission and teaching processes.

Hooshyar and his colleagues have developed two learner modelling approaches in this realm (Hooshyar, 2024; Hooshyar et al., 2024), investigating the potential of this AI family for responsible and trustworthy AI in education. The first incorporates educational causal relationships to guide learning and extract human-understandable rules from the network's predictions (Hooshyar et al., 2024). Their study compared traditional deep neural networks with a knowledge-based deep neural network (NSAI) trained to predict learner performance. The findings revealed that NSAI, which integrates both raw data and symbolic knowledge, exhibited better generalizability compared to traditional models. More importantly, NSAI accurately captured causal relationships, facilitating the extraction of meaningful rules, which allows for refining domain knowledge and enhancing interpretability in predictive modelling. To further assess the reliability of explainable AI methods, they evaluated four widely used explainer algorithms in education for accuracy in representing the "inner workings" of deep neural networks by extracting and comparing knowledge from trained models (Hooshyar & Yang, 2024). The second approach embedded educational knowledge into the loss function of unsupervised deep neural networks, enabling the model to penalize behaviours that deviate from predefined educational principles (Hooshyar, 2024). Their findings highlighted two key concerns: (1) *while deep neural networks often exhibit strong predictive performance, they may depend on spurious correlations, compromising their reliability*; (2) *post-hoc explainer algorithms, essentially additional "black boxes" on top of already opaque deep learning models, may provide explanations that do not accurately reflect the model's decision-making process*. This misalignment makes it difficult to interpret the model's inner workings, which is essential for understanding learners' knowledge development and hidden learning patterns in education, **raising questions about interpretability and trustworthiness**.

Recent regulatory frameworks, such as the EU AI Act (European Union, 2024), represent a significant shift from voluntary ethical guidelines to legally enforceable regulations, introducing stringent requirements for transparency, accountability, fairness, and governance. *Meeting these rigorous standards is particularly challenging without considering hybrid approaches like NSAI, which integrate domain knowledge, reduce biases, enhance explainability, and support human oversight throughout the AI lifecycle*. Despite the proven potential of NSAI and its successes, as well as the urgent need for responsible and trustworthy AI that is ethically sensitive and fair, there remains a lack of studies systematically comparing the AI methods in predictive modelling. Addressing this gap is crucial for advancing responsible, trustworthy, and regulation-compliant AI-driven learning systems without compromising their effectiveness.

## 3. Method

Fig. 1 illustrates the overall architecture of our comparative study, examining the capabilities of different AI method families in educational data mining from both quantitative (generalizability) and qualitative (interpretability) perspectives. We briefly outline the sample and data collection procedure, followed by data pre-processing, model training, validation, and explanation components.

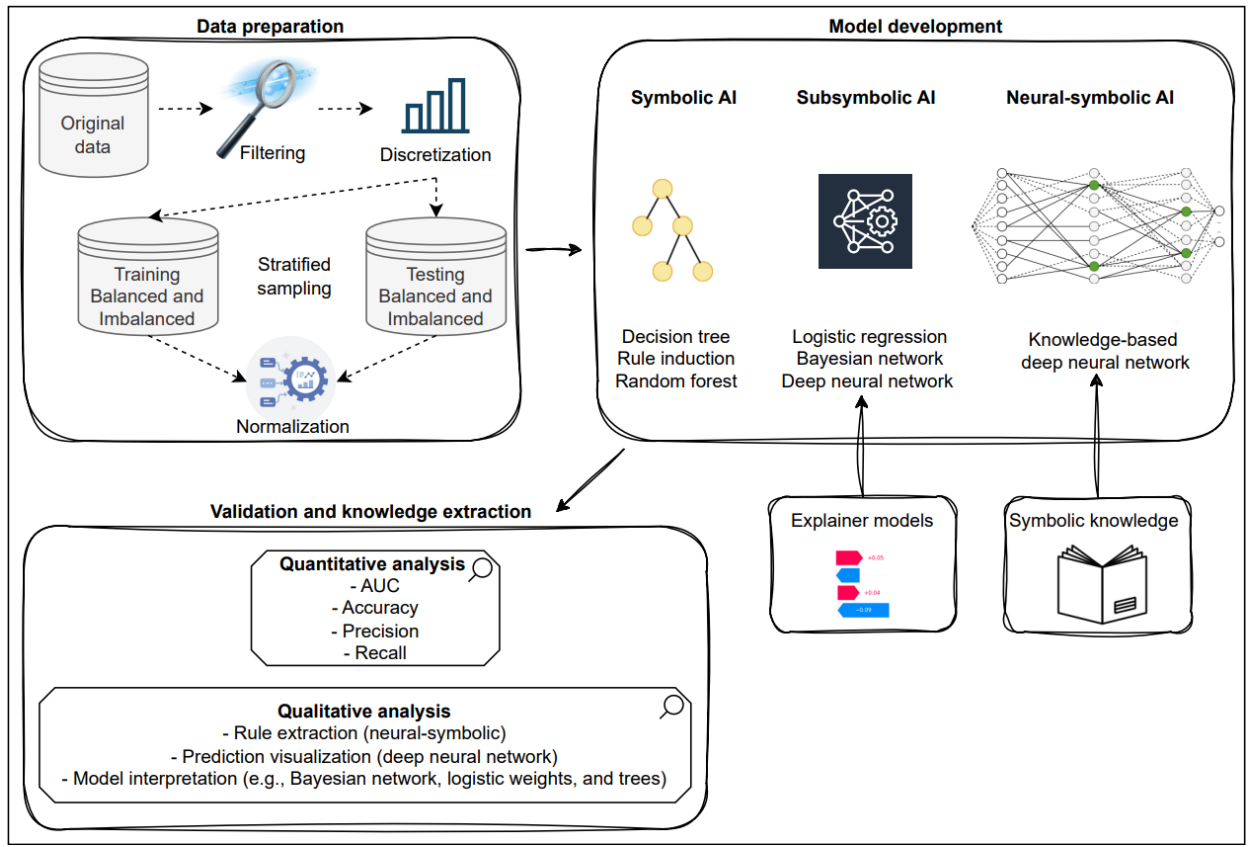

**Fig. 1** Overall architecture of the comparative approach

### 3.1. Sample and procedure

To predict 7th-grade students' performance on the national math test, we used data from the 3rd-grade assessment conducted in Estonian schools with a web-based SRL tool (see Section 1.2 and Kikas (2018)). The participants included 453 students (225 boys) from 33 schools. The assessment of SRL indicators in Grade 3 was organized by the Estonian Education and Youth Board (HARNO) and conducted in computer labs under the supervision of teachers in 2018. Participants' math skills were assessed in 2022 as part of the regular national testing in Estonia. The data from the two assessments were merged by a representative from HARNO. All data were anonymized before being shared with the researchers. Students and their parents were informed about the assessment, and students were given the option to withdraw at any time.

#### 3.1.1. Measures

An overview of the tasks, the indicators they assess across three areas of SRL (motivation, cognition, and metacognition), the acquired knowledge, and the variable names used in the analyses are provided in Table 1. While motivation, cognition, and metacognition area describe self-reported SRL competencies, acquired knowledge shows what students had learnt during completing the test. Below, we briefly describe each task and indicator.

Growth mindset was assessed using four items based on questionnaires by Gunderson et al. (2018), and Schommer-Aikins et al. (2000). Students responded to each statement with "No," "Yes," or "Don't know." The total number of "No" responses was used as the score. Self-efficacy was evaluated using students' self-reported confidence before completing the calculation, word memorization, and word-searching tasks. They selected their response from "No," "Yes," or "Sometimes," and the total number of "Yes" responses was used as the score. Motivation was assessed with a modified version of the self-regulatory style questionnaire by Assor et al. (2009). The instructions explained that children study for different reasons and described several of these reasons. Students rated how well each statement described their own motivation on a five-point scale. Autonomous motivation was assessed with two items (e.g., "I put effort into learning because learning is fun"), while controlled motivation with six items (e.g., "I put effort into learning because I want to avoid feeling ashamed"; "I put effort into learning because I want to avoid punishment"). It is important to note that the SRL competencies were all measured prior to engaging with the math task. They represent students' self-perceptions prior to task engagement, not actual SRL knowledge and skills that were used during the math task.

The math task consisted of 10 calculation problems (e.g., 8 - 1 =?), where each correct answer revealed a piece of a puzzle. Students were instructed to complete the task as quickly as possible. After solving four problems, they were told that time was up and asked whether they wanted to continue solving the remaining problems. Choosing to proceed was coded as 1, while not proceeding was coded as 0. Regardless of their response, all students were redirected to complete the remaining problems. During this phase, they had the option to click the *Help* button to receive the correct answer. The help-seeking score was calculated based on the number of times students clicked the button.

In the categorization task, students were shown four photos of objects (e.g., soap, toothbrush, book, pencil) and four photos of children interacting with these objects (e.g., a girl washing her face with soap, a person reading a book). Students were asked to group the photos into four sets of two and then reorganize them into four new sets of two. Next, they categorized the photos into two groups (four photos per group) and then into another set of two groups. Two scores were calculated: heterogeneous thinking level (sum of all different categories; max: 12) and scientific thinking level (sum of scientific groups; max: 6). A categorization was considered scientific if students grouped items based on abstract categories (e.g., children vs. objects), while an everyday categorization was based on contextual or functional grouping (e.g., a girl washing her face with soap; see Kikas et al. (2008).

In the word searching task, students had to find and mark the word "sinine" among 243 words twice (each within one minute). The attention score was calculated as the total number of correctly identified words across both attempts (max: 54).

In the word memorization task I & II, students were given 90 seconds to memorize 21 Estonian-language nouns from three categories: vegetables (e.g., potato), sports equipment (e.g., ball), and weather phenomena (e.g., snow) (see Hennok et al. (2023) and Kikas & Jõgi (2016)). The words were randomly displayed together on a computer screen, each consisting of 4–6 letters and commonly used in everyday language. Students were then asked to recall and identify as many words as possible from a list of 35 words, selecting up to 21 words. They were neither informed about the categories nor instructed to categorize them. The task was given twice: at the beginning of the test and after the students had read a text on how to memorize words better. Word memorization I and II scores were calculated as the sums of correctly identified words. After completing the word memorization tasks, students estimated how many words they had recalled correctly, which was used as an indicator of self-confidence in learning words I & II. They were then presented with six specific memorization strategies and asked to indicate which they had used. Two strategies involved rehearsal (reading the words several times; repeating the words several times), while four involved association (visualizing objects corresponding to the words and memorizing the visualization; forming sentences from the words and memorizing the sentences; grouping words by their first letter and memorizing the groups; grouping words by meaning and memorizing the groups). Dummy-coded association strategy use I and II was applied (1 = used; 0 = did not use). After reporting strategy use in the first task, students rated the effectiveness of each strategy for memorizing target words. The same six strategies were shown again and evaluated on a five-point scale. These ratings served as indicators of metacognitive awareness of strategies, with composite scores for valuing rehearsal and association strategies. After the second task, students indicated which memorization task they performed better. Self-evaluation was coded as correct (1) or incorrect (0).

After the first memorization task students read a text (273 words, 27 sentences) explaining the most effective memorization methods (Kikas et al., 2021). They had unlimited time to read before answering five multiple-choice comprehension questions with one correct answer. The text comprehension score was the sum of correct answers. After the test, students were shown seven reading strategies (two effective, five ineffective) and asked to select the one they used. Using an effective reading strategy was coded as 1 (used) and 0 (did not use). Next, the same six reading strategies (excluding "I did not read till the end") were presented, and students rated their effectiveness on a five-point scale. A score reflecting the value placed on effective reading strategies was calculated. Finally, the text was shown again, and students were asked to mark up to five sentences they considered most important, indicating their ability to identify key information.

The national math test included 48 tasks, with 31 assessing factual and procedural knowledge and 17 assessing conceptual knowledge. An overview of the test tasks and results for all participating students is available on the HARNO webpage[2].

### 3.2. The comparative approach
### 3.2.1. Data preparation

We started the data pre-processing by filtering the variables that do not fall under the scope of our task (e.g., school name) and filtering examples with missing values. Next, we discretized the labels (conceptual and factual knowledge) into categorical classes. To comprehensively evaluate our models, we created two separate datasets—one balanced and one imbalanced—as different AI methods perform differently depending on the class distribution. In the educational context, identifying low-performing students is crucial for timely interventions, making class imbalance a significant factor to consider. To create the balanced dataset, we used the cut-off mean (since both the mean and median were closely similar), resulting in two labels: Low (denoted 0) and High (denoted 1) performers. For the imbalanced dataset, we used the 1st quartile as a threshold, labelling students as Low performers and Others (hereafter, for simplicity and consistency, we refer to the latter as High). This approach specifically aimed at identifying struggling students. To handle the class imbalance challenge, we chose not to use resampling techniques. Undersampling could reduce the available data in an already small dataset, potentially leading to information loss, while oversampling could introduce redundancy, distorting the real data distribution, leading to issues such as small disjuncts, class overlap, and dataset shift (Fernández et al., 2018). Instead, we used the original natural dataset, allowing models to learn from the true data distribution and providing an unbiased evaluation of how well different AI methods handle class imbalance without introducing synthetic biases.

Since measuring model performance on a separate test set is essential to prevent poorly performing models from misidentifying patterns and leading to misleading interpretations, we used stratified sampling to split our data into training and test sets. This approach ensures a fair representation of each class and minimizes bias in the evaluation process. Finally, we applied min-max scaling to normalize the data within the range [0,1], ensuring uniformity across variables that were originally on different scales. This step was particularly important because some machine learning methods, such as deep neural networks, are sensitive to scale variations, which can affect convergence and performance. To avoid data leakage, normalization was performed after splitting the datasets.

---

[2] https://projektid.edu.ee/pages/viewpage.action?pageId=184849361

### 3.2.2. Model development

In general, AI methods can be categorized differently based on focus (Bader & Hitzler, 2005; Sarker et al., 2021). When considering explainability, symbolic AI would include methods such as rule-based models, decision trees, Bayesian networks, and even logistic regression, while sub-symbolic AI would encompass models such as random forests and neural networks, due to their complex decision-making. From a knowledge representation perspective, however, decision trees, rule induction, and even random forests can be considered as symbolic, whereas Bayesian networks, logistic regression, and neural networks would fall under sub-symbolic AI. Although the Bayesian networks and random forests could as well be considered hybrid, due to combining symbolic structures with statistical inference. This study adopts the knowledge representation perspective for comparison. Based on this classification, we trained and evaluated the models from each AI family using 10-fold cross-validation. The models were selected to represent key symbolic, sub-symbolic, and hybrid approaches, ensuring a comprehensive comparison.

For the symbolic AI family, we trained a decision tree (Lewis, 2000) using default parameters of the gain ratio criterion, with a maximum depth of 10 and post-pruning set at a confidence level of 0.1. The rule induction (Cohen, 1995) was also trained using the information gain criterion, with a sample ratio of 0.9, a pureness threshold of 0.9, and a minimal pure benefit of 0.25. For random forest (Breiman, 2001), we used the gain ratio criterion, trained with 100 trees, a confidence voting strategy, and post-pruning with a confidence level of 0.1. For the sub-symbolic AI family, we implemented kernel logistic regression (Keerthi et al., 2005) as it efficiently handles complex decision boundaries using kernel methods, unlike standard logistic regression, which assumes linear separability and may struggle with non-linear patterns. We used default parameters, including kernel type = dot, kernel cache = 200, C = 1.0, convergence epsilon = 0.001, and a maximum of 100,000 iterations. Additionally, we trained a Bayesian network using the PC structure learning method (Spirtes & Glymour, 1991), setting the significance threshold to 0.07, which was determined experimentally. Before learning, we provided background knowledge, including that no variable influences gender, but gender influences others. No arrows would originate from factual and procedural knowledge as they serve as labels. We also restricted arrows from variables with the suffix 2 (e.g., MetaWordev2, LNmemwords2) to those with the suffix 1, as the latter represent prerequisites. For the deep neural network (Chollet, 2017), we used a three-layer architecture with 64, 32, and 1 neurons, ReLU activation for the hidden layers, sigmoid activation for the output layer, and binary cross-entropy loss function. The model was optimized using Adam with a learning rate of 0.001, trained over 100 epochs with a batch size of 32, and implemented using the Keras library.

For the neural-symbolic AI family, we implemented knowledge-based artificial neural networks (KBANN) (Towell & Shavlik, 1994), which integrate predefined symbolic rules into the neural network's structure. This approach enables the network not only to learn from data but also to leverage domain-specific symbolic knowledge, improving interpretability. The process begins by loading and adapting symbolic rules, which guide the network's architecture, layer configuration, weight initialization, and connections. The model is trained using a modified backpropagation algorithm, ensuring that learning remains aligned with the embedded knowledge while adjusting network parameters. After training, the model extracts and reconstructs rules, allowing it to generate explanations that align with the original symbolic knowledge, thereby improving transparency. We used default KBANN parameters, including a learning rate of 0.001, gradient descent optimizer, and binary cross-entropy loss function. Additionally, a dropout mechanism was applied during training to enhance generalization. The model underwent iterative refinement, ensuring that weights and biases remained aligned with predefined symbolic rules, thereby preserving interpretability while optimizing accuracy.

### 3.2.3. Validation and knowledge extraction

To comprehensively validate the models, we compared their performances both quantitatively and qualitatively. For quantitative evaluation, we used accuracy, area under the curve (AUC), recall, and precision. Accuracy provides a general measure of overall performance, while precision and recall offer insights into the models' ability to correctly classify instances of each class. AUC measures the model's ability to distinguish between classes across different classification thresholds, providing a more comprehensive evaluation of classification performance, especially in imbalanced datasets. For qualitative analysis, we compared the interpretability of the learned models. As decision trees, rule induction, random forest, Bayesian networks, and kernel logistic regression provide built-in explanations for their models, we only applied SHAP kernel explainer to provide global and local explanations for the deep neural network model. The NSAI method of KBANN offers explanations by extracting symbolic rules that align with the injected domain knowledge, ensuring that the model remains interpretable and its predictions are aligned with human-understandable reasoning.

## 4. Results and Analysis
### 4.1. Dataset and educational knowledge

Table 1 provides the descriptive statistics of the dataset's features, while Figures 2a and 2b illustrate the distribution of factual and conceptual knowledge datasets using deviation charts.

**Table 1** Descriptive statistics of the original dataset

| Task | Indicator | SRL area | Variables | Min | Max |
|---|---|---|---|---|---|
| | | | Gender | 0 | 1 |
| Questionnaire | Growth mindset | Mot | MotiBeliefs | 0 | 4 |
| Questionnaire | Autonomous motivation | Mot | MotiInterest | 1 | 5 |
| Questionnaire | Controlled motivation | Mot | MotiKontrMot | 1 | 5 |
| Math task | Help seeking | Mot | MotiMathHelp | 0 | 12 |
| Math task | Proceeding with calculation | Mot | MotiProceed | 0 | 1 |
| Questions | Self-efficacy | Mot | MotiSEF | 0 | 3 |
| Word searching task | Attention | Cog | CogAttention | 0 | 43 |
| Categorization task | Heterogeneous thinking level | Cog | CogHeterogeneous | 0 | 12 |
| Categorization task | Scientific thinking level | Cog | CogScientific | 0 | 6 |
| Word memorizing task I | Association strategy use I | Cog | CogMemorystrategies1 | 0 | 1 |
| Word memorizing task I | Memorized words I | Cog | CogMemwords1 | 0 | 21 |
| Reading task | Using effective reading strategy | Cog | CogReadstrat | 0 | 1 |
| Reading task | Identifying important information | Cog | CogTextinfo | 0 | 5 |
| Word memorizing task I | Valuing rehearsal strategies | Meta | MetaMemRehearsal | 1 | 5 |
| Word memorizing task I | Valuing association strategies | Meta | MetaMemstratAssoc | 1 | 5 |
| Reading task | Valuing effective reading strategies | Meta | MetaReadstrat | 0 | 8 |
| Word memorizing task II | Self-evaluation | Meta | MetaWord12ev | 0 | 1 |
| Word memorizing task I | Self-confidence in learning words I | Meta | MetaWordev1 | 1 | 3 |
| Word memorizing task II | Self-confidence in learning words II | Meta | MetaWordev2 | 1 | 3 |
| Word memorizing task II | Association strategy use II | Learnt knowledge | LNmemorystrategies2 | 0 | 1 |
| Word memorizing task II | Memorized words II | Learnt knowledge | LNmemwords2 | 0 | 21 |
| Text comprehension test | Text comprehension | Learnt knowledge | LNtext | 0 | 5 |
| National math test | Factual math knowledge | Outcome | MF | 2 | 31 |
| | Conceptual math knowledge | Outcome | MC | 0 | 17 |
| National math test | Factual math knowledge | Outcome | MF (label) | Balanced* = 0s: 205, 1s: 248 Imbalanced** = 0s: 125, 1s: 328 | |
| | Conceptual math knowledge | Outcome | MC (label) | Balanced = 0s: 223, 1s: 230 Imbalanced*** = 0s: 125, 1s: 328 | |

Note. Mot = motivational area; Cog = cognitive area; Metacog = metacognitive area; * Discretized based on average;
** Discretized based on 1st quartile = 14; *** Discretized based on 1st quartile = 6

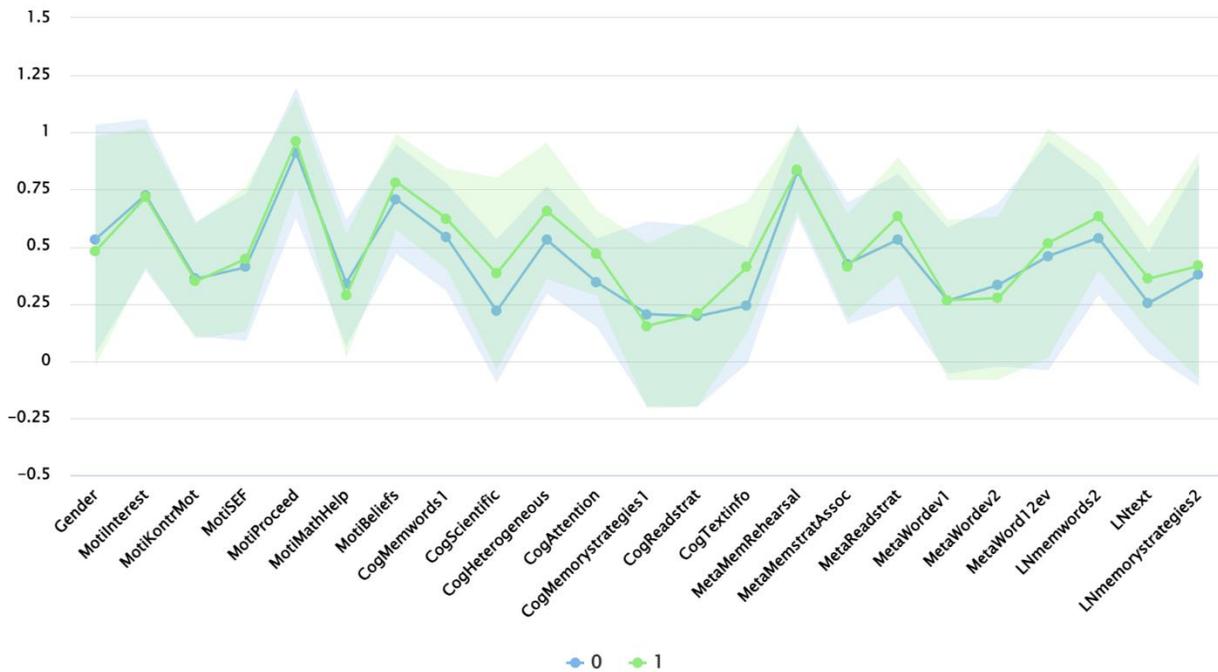

(a)

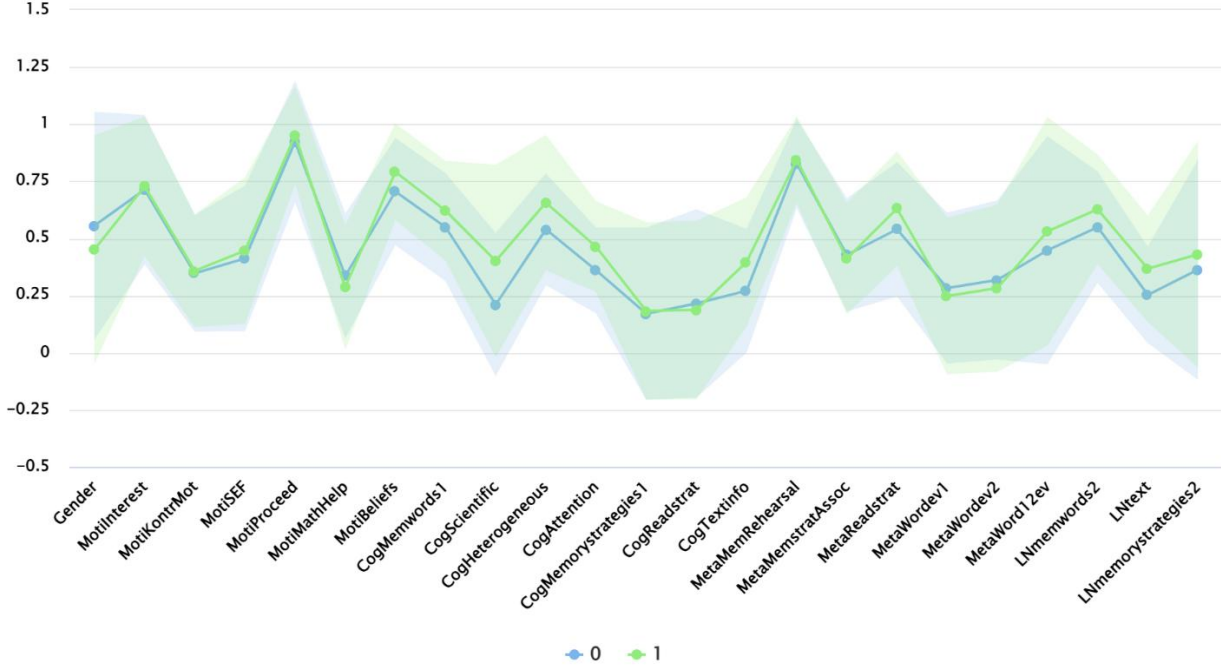

(b)

**Fig. 2** Visualization of distribution of the: a) factual knowledge (MF), and b) conceptual knowledge (MC) datasets, where 0 and 1 denote low and high performers, respectively

To incorporate domain knowledge into the NSAI approach, we adapted the SRL framework proposed by Kikas et al. (2024). This framework emphasizes that learners' performance is influenced by their self-regulatory behaviours throughout the learning process, particularly in cognition (both domain-general and domain-specific), metacognition, motivation, and emotional factors. To represent the knowledge within neural networks, we utilized propositional logic, specifically a set of non-recursive Horn clauses, as illustrated below.

> *$MF^3$: - Metacognition, Cognition, Motivation, LearntKnowledge.* (R1)
> *Metacognition: - MetaMemRehearsal, MetaMemstratAssoc, MetaReadstrat,*
>   *MetaWord12ev, MetaWordev1, MetaWordev2.* (R2)
> *Cognition: - CogAttention, CogHeterogeneous, CogMemorystrategies1,*
>   *CogMemwords1, CogReadstrat, CogScientific, CogTextinfo.* (R3)
> *Motivation: - MotiBeliefs, MotiInterest, MotiKontrMot, MotiMathHelp,*
>   *MotiProceed, MotiSEF.* (R4)
> *LearntKnowledge: - LNmemorystrategies2, LNmemwords2, LNtext.* (R5)

In summary, the first propositional logic rule (R1) suggests that final performance or scores are influenced by metacognition, cognition, motivation, and learnt knowledge (which represents test-specific cognition). Since no emotion-related data was collected, the emotion component is absent. The remaining rules (R2-R5) then align with our framework and corresponding variables. For example, cognition (R3) is represented by variable attention, heterogeneous thinking level, scientific thinking level, association strategy use I, memorized words I, using effective reading strategy, and identifying important information.

---

[3] The same rule structure applies to MC, where MF is replaced with MC, maintaining the hierarchical relationships among metacognition, cognition, motivation, and learnt knowledge.

### 4.2. Experiment one: Balanced data

This experiment assesses the performance of symbolic, sub-symbolic, and neural-symbolic AI models under balanced data conditions, where low and high performers are equally represented. This allows us to establish a baseline for model performance across factual and conceptual knowledge classification, comparing them both quantitatively and qualitatively.

#### 4.2.1. Quantitative comparison

Table 2 presents the performance of various models across different evaluation metrics. While some models achieve high AUC or precision for factual knowledge, NSAI appears to be among the strongest overall performers in conceptual knowledge while remaining competitive in factual knowledge. It achieves the highest recall for low performers while maintaining acceptable recall for high performers at the standard classification threshold of 0.5, commonly used in AI for education and educational data mining research (Hooshyar et al., 2022; Romero & Ventura, 2013). This threshold ensures comparability across studies, prevents arbitrary bias towards one class, and maintains a standardized decision boundary to reduce the risk of inconsistent interpretations. Symbolic and sub-symbolic models generally struggle to identify enough at-risk students, facing trade-offs between precision and recall. By integrating symbolic knowledge, NSAI seems to mitigate these limitations, leveraging structured domain expertise to enhance recall without significantly compromising precision. As a result, NSAI maintains a well-balanced performance in predicting both factual and conceptual mathematical knowledge while improving the identification of students needing support.

**Table 2** Performance of various AI methods in predicting both factual and conceptual knowledge in balanced unseen test dataset

| AI family | methods | Metrics | Factual knowledge (MF) | | Conceptual knowledge (MC) | |
|---|---|---|---|---|---|---|
| | | | Low Performer | High Performer | Low Performer | High Performer |
| Symbolic | Decision tree | AUC (%) | 53 | | 53 | |
| | | Accuracy (%) | 63 | | 55 | |
| | | Precision (%) | 61 | 63 | 56 | 55 |
| | | Recall (%) | 46 | 76 | 53 | 57 |
| | Rule induction | AUC (%) | 65 | | 51 | |
| | | Accuracy (%) | 67 | | 56 | |
| | | Precision (%) | 70 | 65 | 56 | 56 |
| | | Recall (%) | 46 | 84 | 49 | 63 |
| | Random forest | AUC (%) | 67 | | 68 | |
| | | Accuracy (%) | 63 | | 64 | |
| | | Precision (%) | 60 | 65 | 63 | 63 |
| | | Recall (%) | 56 | 70 | 62 | 65 |
| Sub-symbolic | Logistic regression | AUC (%) | 70 | | 70 | |
| | | Accuracy (%) | 64 | | 60 | |
| | | Precision (%) | 61 | 71 | 61 | 69 |
| | | Recall (%) | 54 | 48 | 56 | 54 |
| | Bayesian network | AUC (%) | 70 | | 64 | |
| | | Accuracy (%) | 67 | | 56 | |
| | | Precision (%) | 74 | 65 | 54 | 59 |
| | | Recall (%) | 41 | 88 | 69 | 43 |
| | | AUC (%) | 64 | | 67 | |

|  |  | Accuracy (%) | 64 || 64 ||
|  | **Neural network** | Precision (%) | 59 | 69 | 63 | 64 |
|  |  | Recall (%) | 66 | 62 | 64 | 63 |
| **Neural-symbolic** | **NSAI** | AUC (%) | 66 || 69 ||
|  |  | Accuracy (%) | 66 || 69 ||
|  |  | Precision (%) | 60 | 72 | 69 | 70 |
|  |  | Recall (%) | 71 | 62 | 69 | 70 |

*Symbolic models.* In *factual knowledge*, symbolic models exhibit varying levels of discriminative ability. Decision trees have a weak AUC of 53%, while rule induction improves to 65%, and random forests perform best with an AUC of 67%, suggesting better differentiation between performance levels. However, despite these AUC values, their recall remains low at the standard threshold, limiting their ability to detect struggling students. While they may be able to identify more low performers by adjusting the threshold, this comes at the cost of higher false positives, leading to increased misclassifications and reduced precision. Decision trees achieve only 46% recall for low performers while over-predicting high performers (76%), making them unreliable for real-world interventions. Rule induction improves precision for low performers (70%) but still suffers from poor recall (46%). While random forests offer slightly better recall (56%), they still fail to identify enough at-risk students. In *conceptual knowledge*, performance declines further, with decision trees achieving only 55% accuracy and 53% AUC, while rule induction achieves 56% accuracy and 51% AUC. Random forests perform best, with a recall of 62%, higher accuracy, and a better AUC, making them the strongest among symbolic models.

*Sub-symbolic models.* For *factual knowledge*, logistic regression and Bayesian networks both achieve an AUC of 70%, indicating a decent ability to distinguish performance levels. However, despite this, logistic regression struggles with recall, identifying only 54% of low performers, while Bayesian networks perform even worse, with a recall of just 41%, failing to detect most at-risk students. Logistic regression maintains strong precision (61% for low and 71% for high performers) and achieves 64% accuracy, whereas Bayesian networks improve accuracy (67%) and precision (74% for low performers) but at the cost of recall. Neural networks achieve the highest recall (66%), capturing more low performers than other models, but their lower precision (59%) increases false positives, potentially leading to unnecessary interventions. In *conceptual knowledge*, logistic regression again reaches a strong AUC (70%) with 60% accuracy but still struggles with recall (56%). Bayesian networks, with an AUC of 64% and 56% accuracy, achieve the highest recall (69%) among sub-symbolic models but suffer from weak precision (54%), increasing false positives. Neural networks provide a better balance, with 64% recall and improved precision (63%), making them a stronger option when prioritizing both early identification and minimizing misclassifications in educational settings.

*Neural-symbolic model.* NSAI demonstrates strong performance across both factual and conceptual knowledge, particularly excelling in recall, a key metric for identifying struggling students. In factual knowledge, it achieves an AUC of 66%, indicating a solid ability to distinguish performance levels. Despite this, it maintains the highest recall (71%), identifying more struggling students than any other model while keeping precision competitive (60%). Its accuracy (66%) further supports its generalizability across performance groups. Unlike symbolic and sub-symbolic models, which often face trade-offs between recall and precision, NSAI effectively balances both by leveraging the strengths of both paradigms. For conceptual knowledge, NSAI continues to perform well, achieving an AUC of 69%, alongside high recall (69%) and accuracy (69%). This strong recall ensures struggling students are detected, while its precision (69%) keeps false positives at a reasonable level. These results suggest that NSAI offers a well-balanced approach for targeted interventions, making it a promising model for improving student outcomes.

### 4.2.2. Qualitative comparison

This section qualitatively interprets the patterns identified by the models for the factual knowledge label (see Appendix A for conceptual knowledge models). Symbolic models primarily emphasize cognitive factors like attention and word memorization, along with motivational aspects such as growth mindset, while incorporating a mix of other (meta)cognitive and motivational variables. In contrast, sub-symbolic models rely mainly on gender, followed by cognitive factors like heterogeneous thinking and attention, with less focus on metacognitive and motivational aspects. The neural-symbolic model, however, adopts a more holistic approach, integrating learnt knowledge, metacognition, motivation, and cognition. Notably, the motivational factors proceeding with calculation and growth mindset, as well as the metacognitive factor valuing rehearsal strategies, followed by autonomous motivation and valuing effective reading strategies, have made a significant contribution to learnt knowledge, highlighting the key role of these motivational and metacognitive aspects in acquiring math factual knowledge. Unlike symbolic and sub-symbolic models, which either prioritize cognitive factors or depend on demographic variables like gender, the neural-symbolic model provides a more interpretable and balanced framework for knowledge acquisition.

*Symbolic Models.* Figure 3 shows the decision tree learned from the training data for predicting factual knowledge. The most generalizable patterns identified in the tree highlight that word memorizing I is the most important predictor for distinguishing low performers (0) from high performers (1). If a student's score on this variable is very low ($\leq 0.024$), they are immediately classified as a low performer. Moving further down the tree, factors such as valuing rehearsal strategies and attention refine the classification. Low scores on these features strongly indicate a low performer classification. To further explain the learned tree, we highlighted one of the branches with high purity. The highlighted branch represents a classification where 38 out of 39 students were correctly classified as high performers (1), demonstrating a highly confident classification. This suggests that students on this path exhibit strong cognitive and metacognitive skills, leading to a reliable prediction of high performance.

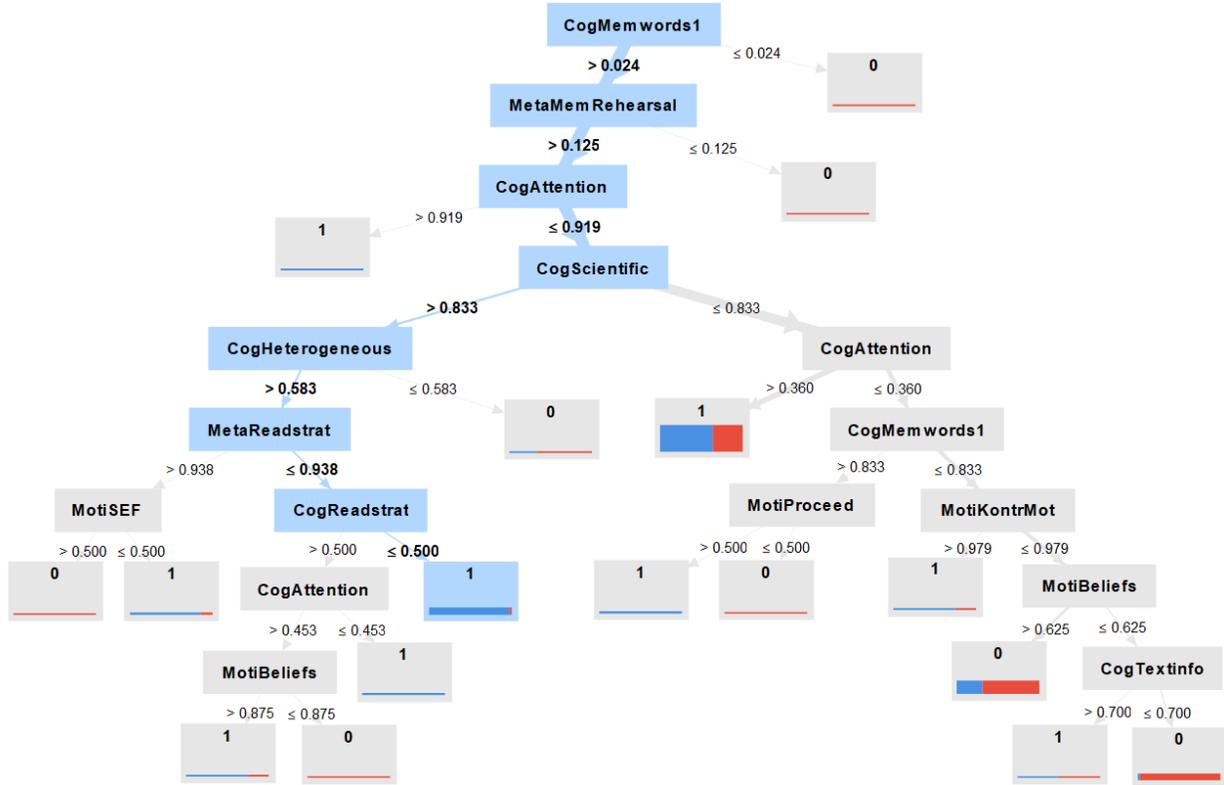

**Fig. 3** The decision tree learned for predicting factual knowledge (MF) using the balanced dataset

Table 3 presents the rules learned from rule induction for predicting factual knowledge. The rules highlight cognitive attention, memory skills, and motivational factors such as growth mindset and help-seeking as key in distinguishing low and high performers. Low performers (0) are primarily identified by low cognitive attention, weak skills in identifying important information from text and text comprehension, and insufficient memory skills, both as indicators of cognition and learnt knowledge areas (*CogMemWords1*, *LNmemWords2*), often combined with lower motivation for seeking help. Notably, high autonomous motivation alone does not guarantee high performance if cognitive skills are lacking. High performers (1) are characterized by strong cognitive attention, better memory, and a growth mindset and help-seeking behaviour, with some rules achieving perfect classification. Overall, these findings suggest that cognition and motivation must work together, reinforcing the need for interventions that address both aspects to improve student outcomes.

**Table 3** Rules learned for predicting factual knowledge (MF) using the balanced dataset

| |
|---|
| if CogAttention > 0.360 and CogHeterogeneous > 0.583 and LNmemwords2 > 0.548 and CogMemwords1 ≤ 0.976 and MotiKontrMot ≤ 0.604 then 1 (59 / 6) |
| if CogAttention ≤ 0.360 and LNtext ≤ 0.100 then 0 (3 / 32) |
| if CogAttention > 0.453 and Gender ≤ 0.500 and CogTextinfo > 0.300 and MetaMemstratAssoc ≤ 0.531 then 1 (16 / 0) |
| if CogTextinfo ≤ 0.100 and CogMemwords1 ≤ 0.690 and CogAttention ≤ 0.291 then 0 (1 / 18) |
| if MotiBeliefs > 0.625 and MotiMathHelp ≤ 0.208 and LNtext > 0.300 and MetaWordev2 > 0.250 then 1 (14 / 0) |
| if MotiInterest > 0.812 and CogAttention ≤ 0.477 and MotiBeliefs ≤ 0.875 and MotiMathHelp > 0.458 then 0 (3 / 16) |
| if CogScientific > 0.833 then 1 (18 / 3) |
| if MetaReadstrat ≤ 0.562 and MotiMathHelp ≤ 0.542 and MetaWordev2 > 0.250 and CogMemwords1 > 0.595 then 0 (0 / 18) |
| if MotiMathHelp > 0.292 and CogReadstrat ≤ 0.500 and MotiInterest > 0.812 and MotiBeliefs ≤ 0.875 then 1 (11 / 0) |
| if CogMemwords1 > 0.690 and CogTextinfo ≤ 0.100 and MotiMathHelp > 0.250 then 1 (7 / 1) |
| if MotiInterest > 0.812 and LNmemwords2 > 0.738 then 0 (0 / 13) |
| else 1 (59 / 51) |
| correct: 281 out of 349 training examples |

Figure 4 presents the pattern highlighted by the random forest model for predicting factual knowledge. The most influential feature is controlled motivation, which has the highest weight, followed by two other motivational factors—help-seeking and autonomous motivation—and two cognitive factors—attention and memorized words I—which constitute the top five most influential factors in predicting factual knowledge. Conversely, features such as association strategy use I, self-evaluation, and the use of effective reading strategies have the lowest impact on factual knowledge predictions. The overall factor ranking highlights that motivational and cognitive factors dominate the model's predictions, while metacognitive and learnt knowledge attributes contribute less significantly.

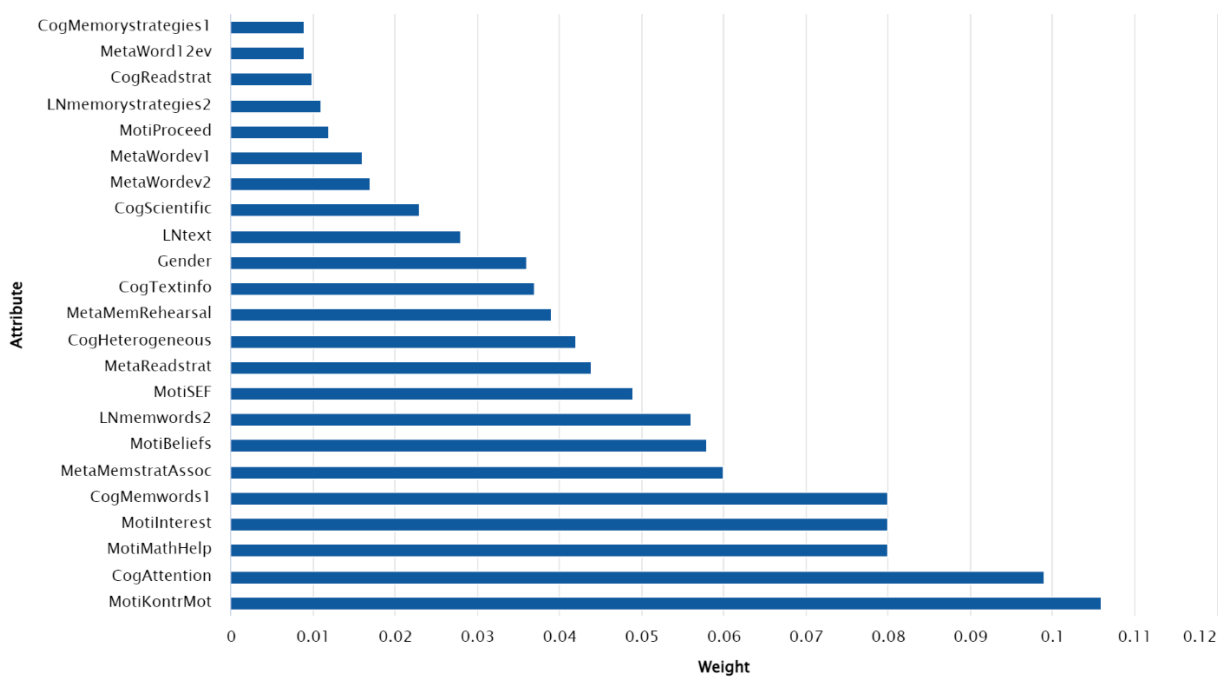

**Fig. 4** Feature importance of the learned random forest for predicting factual knowledge (MF) using the balanced dataset

*Sub-symbolic models.* Figure 5 presents the pattern learned from logistic regression for predicting factual knowledge. Unlike standard logistic regression models that provide explicit coefficients for each feature, this visualization represents feature importance learned through a support vector machine with kernel logistic regression. The key difference is that traditional logistic regression assigns direct coefficients that indicate the impact of each variable on the log-odds of the outcome, whereas this model leverages kernel-based transformations, mapping data into a higher-dimensional space to improve classification. As a result, the weights shown in the figure reflect the influence of each attribute on the decision boundary rather than simple linear coefficients. The positive weights (e.g., gender) suggest an increased likelihood of being classified as high performers, whereas negative weights (e.g., heterogeneous thinking level, attention) indicate a higher likelihood of being classified as low performers, highlighting the influence of cognitive and motivational factors on factual knowledge acquisition.

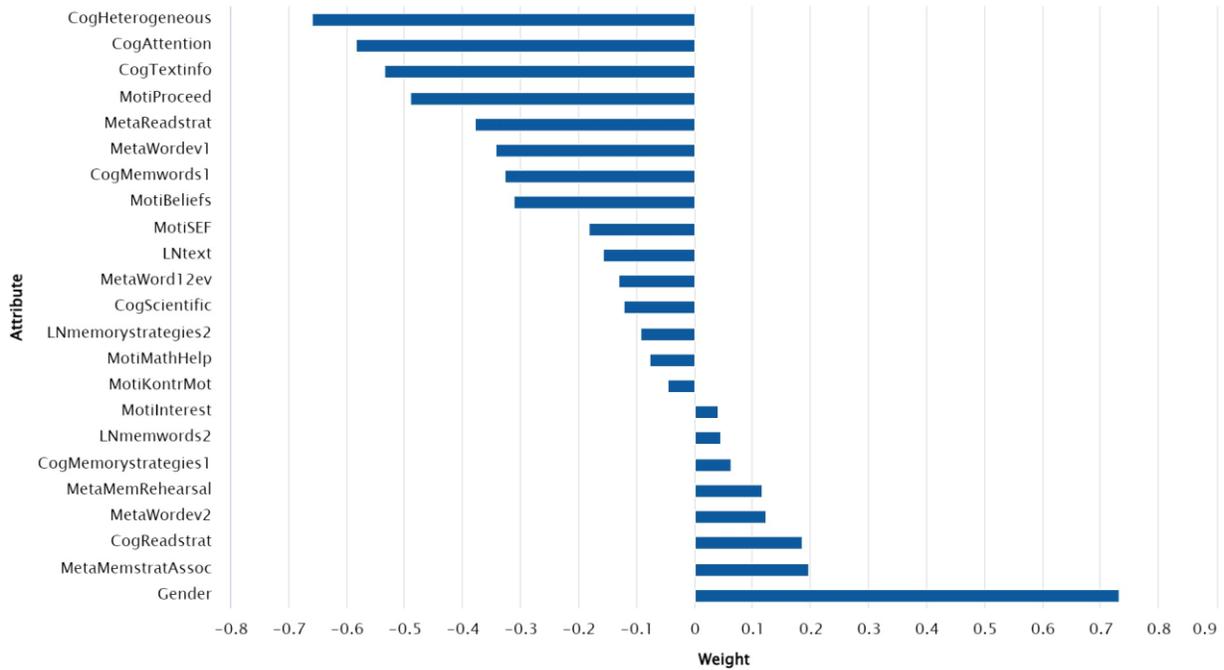

**Fig. 5** The attribute importance based on the learned logistic regression model learned for predicting factual knowledge (MF) using the balanced dataset

Figure 6 illustrates the pattern learned from the Bayesian network model for predicting factual knowledge, highlighting key dependencies between cognitive, metacognitive, and motivational factors. Table 4 presents the top 10 strongest influence factors, providing insights into how these variables shape factual knowledge. Attention and heterogeneous thinking level from the cognitive domain, along with growth mindset, are the strongest predictors of math factual knowledge, with influence scores of 0.75, 0.73, and 0.67, respectively. Beyond these direct predictors, self-confidence in learning words II, memorized words I, identifying important information, heterogeneous thinking level, and using effective reading strategies strongly impact memorized words II. Additionally, heterogeneous thinking level has a strong influence on scientific thinking level, while proceeding with calculation significantly affects text comprehension, ranking among the top 10 strongest connections.

**Table 4** Top 10 influence strengths ranked in descending order

| Parent | Child | Maximum |
| --- | --- | --- |
| MetaWordev2 | LNmemwords2 | 0.83 |
| CogAttention | MF | 0.75 |
| CogMemwords1 | LNmemwords2 | 0.74 |
| CogTextinfo | LNmemwords2 | 0.74 |
| CogHeterogeneous | MF | 0.73 |
| CogHeterogeneous | LNmemwords2 | 0.72 |
| CogHeterogeneous | CogScientific | 0.67 |
| MotiBeliefs | MF | 0.67 |

| MotiProceed | LNtext | 0.60 |
| CogReadstrat | LNmemwords2 | 0.57 |

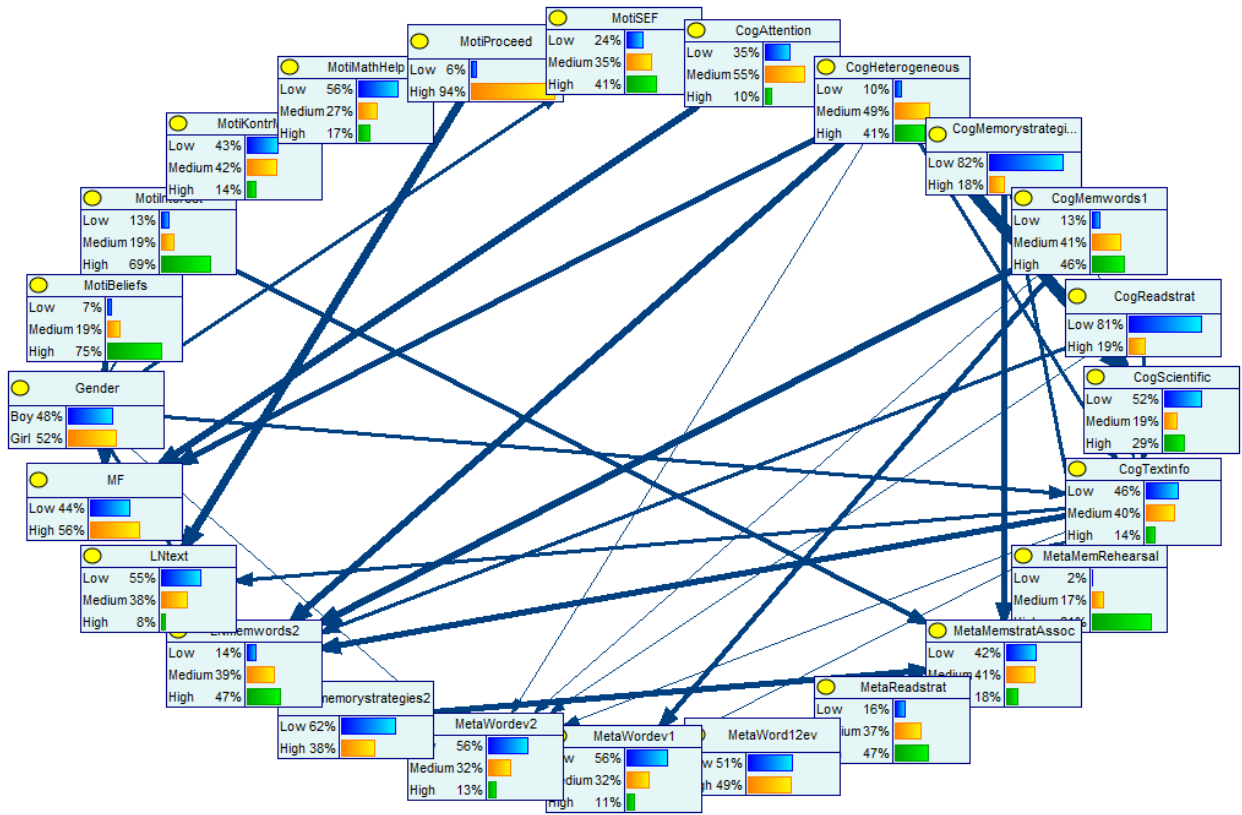

**Fig. 6** The Bayesian network learned for predicting factual knowledge (MF) using the balanced dataset

Figure 7 presents the pattern highlighted by the SHAP global explanation of neural networks for predicting factual knowledge. The plot visualizes the impact of each feature on the model's predictions, with positive SHAP values pushing the prediction towards high performers and negative values towards low performers. Features appearing higher on the plot, such as gender, identifying important information, and heterogeneous thinking level, have a greater overall influence on the model compared to those lower on the list. Additionally, higher feature values (red) tend to contribute positively to the prediction, while lower values (blue) often have a negative impact, reflecting complex interactions between cognitive, metacognitive, and motivational indicators and learnt knowledge in factual knowledge prediction. For example, the first line shows that lower values (0 represents male and 1 represents female in the original dataset) tend to contribute positively, whereas higher values contribute negatively, suggesting that male students are more likely to be classified as high performers in factual knowledge.

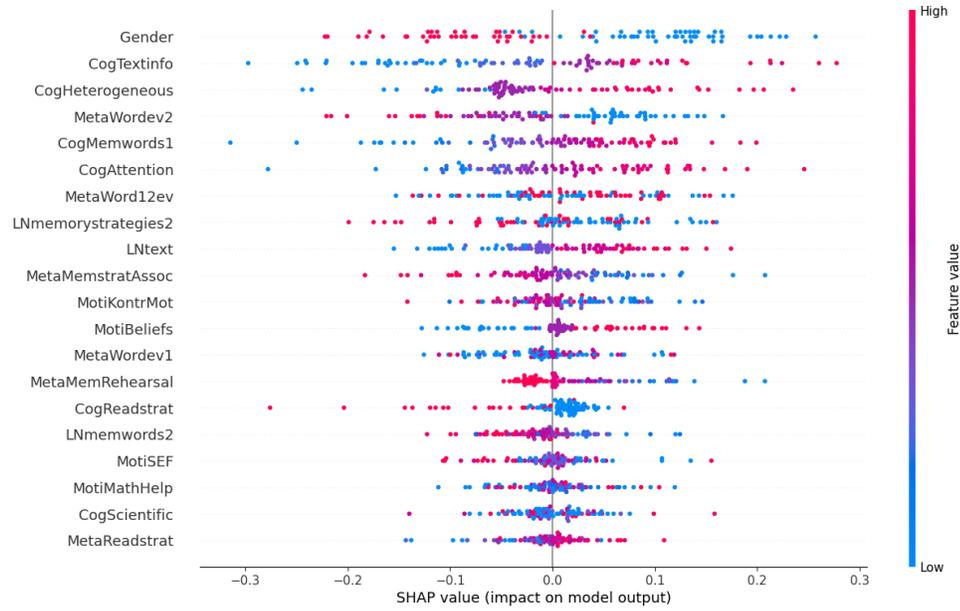

**Fig. 7** SHAP global explanation of the learned neural network for predicting factual knowledge (MF) using the balanced dataset

Figure 8a and 8b illustrate the pattern highlighted by the SHAP local explanation of neural networks for predicting factual knowledge. According to Figure 8a, which presents a correctly classified high performer from the test data, the model assigned a high probability to the correct classification. Features such as gender (+0.13), valuing rehearsal strategies (+0.10), and autonomous motivation (+0.08) contributed the most to this prediction, pushing the final score towards 1.0. The positive contributions (red) from cognitive and motivational attributes outweigh the small negative influences (blue), leading to a confident and correct classification. In contrast, Figure 8b represents a misclassified example, where the model incorrectly predicted a low performer as a high performer. Here, Gender (+0.19), controlled motivation (+0.14), and association strategy use II (+0.14) played a significant role in pushing the prediction towards 1.0, even though some features, such as heterogeneous thinking level (-0.05) and memorized words II (-0.04), indicated a lower likelihood of high performance. Despite the presence of negatively contributing attributes, the overall sum of positive contributions dominated, leading to an incorrect classification. This misclassification suggests that certain features, particularly gender and motivational strategies, may have an outsized influence, potentially leading to overconfidence in incorrect predictions.

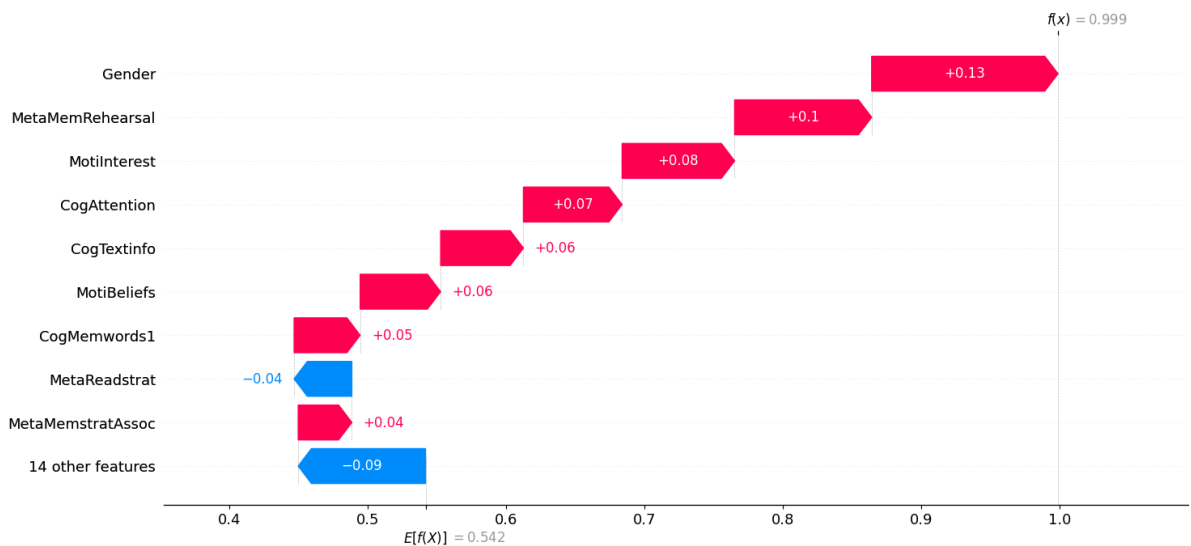

(a)

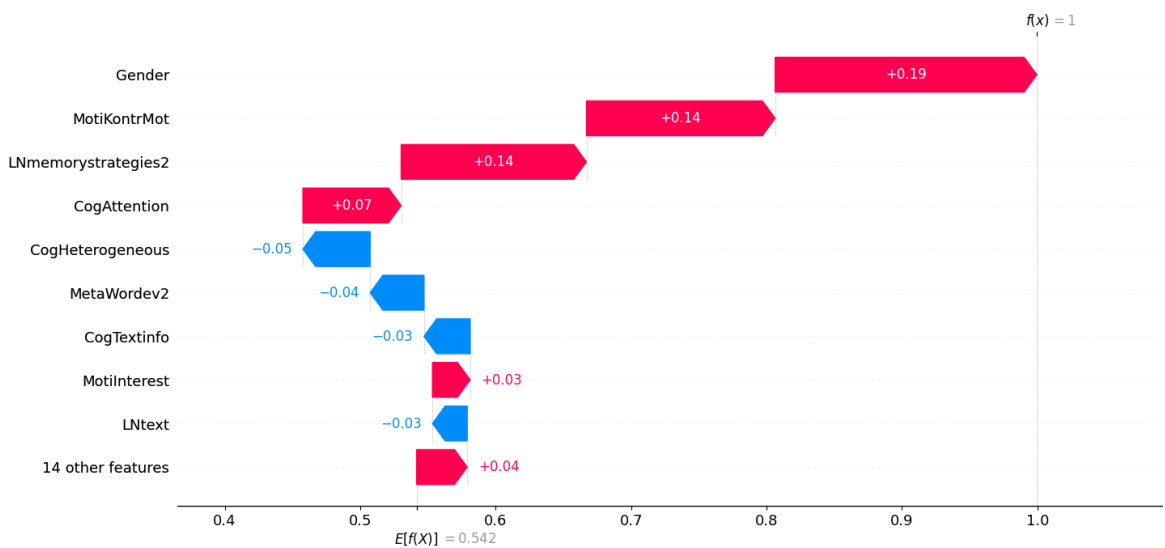

(b)

**Fig. 8** SHAP local explanation of the learned neural network for predicting factual knowledge (MF) using the balanced dataset: a) correctly classified as high performer, and b) misclassified as high performer

*Neural-symbolic model.* Fig. 9a and 9b illustrate the high-level and low-level (more detailed) interpretation of knowledge extracted from the NSAI model for predicting factual knowledge. As seen in Fig. 9a, the most influential latent variable is learnt knowledge, followed by metacognition, motivation, and cognition, indicating their crucial role in factual knowledge acquisition. The variable head1 has the least weight, suggesting a minor contribution.

Fig. 9b provides a detailed breakdown of how different observed variables contribute to these latent constructs. Cognitive attributes such as attention, heterogeneous thinking level, association strategy use I, word memorization I, using effective reading strategies, and scientific thinking level strongly load onto cognition, highlighting their essential role in shaping cognitive processes. Similarly, motivational, metacognitive, and learnt knowledge-related attributes contribute to their respective constructs, emphasizing their importance in factual knowledge acquisition. Interestingly, the motivational factors proceeding with calculation and growth mindset, as well as the metacognitive factor valuing rehearsal strategies, followed by autonomous motivation and valuing effective reading strategies, have made a significant contribution to learnt knowledge, highlighting the key role of these motivational and metacognitive aspects in acquiring math factual knowledge. Although head1 has the lowest overall weight, its highest contributing features include attention, heterogeneous thinking level, word memorization I, scientific thinking level, identifying important information, text comprehension, and valuing effective reading strategies. These features suggest that head1 is linked to a combination of attention, heterogeneous thinking level in cognitive processing, memory-based word retrieval, scientific thinking level, identifying important information during learning, text comprehension, and valuing effective reading strategies. The presence of both cognitive (e.g., attention, memory, and scientific thinking) and metacognitive (e.g., valuing effective reading strategies) elements indicates that head1 captures aspects of how learners engage with, process, and regulate textual and scientific information in factual knowledge acquisition. Given these contributing features, head1 could be labelled as *thinking level and information processing*. This name reflects its role in encoding reading comprehension, structured reasoning, and cognitive control over scientific and text-based information, which, while not the dominant predictor, still plays a distinct role in learning.

Overall, considering the global feature importance in Fig. 10, which shows that motivational and metacognitive factors (constituting six of the top ten) outweigh cognitive and learnt knowledge factors (only four), it becomes evident that besides cognition, motivational and metacognitive aspects play a crucial role in learning outcomes. The extracted NSAI knowledge rules align with these insights, demonstrating how factual knowledge acquisition is shaped by a complex interplay of motivational, metacognitive, cognitive, and learning-based knowledge, further enhancing the model's interpretability.

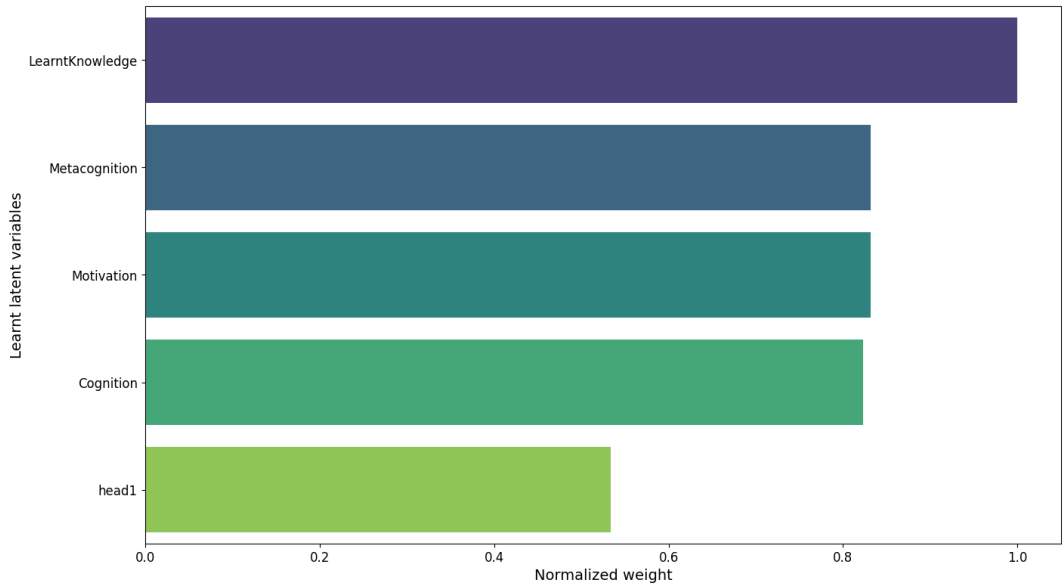

(a)

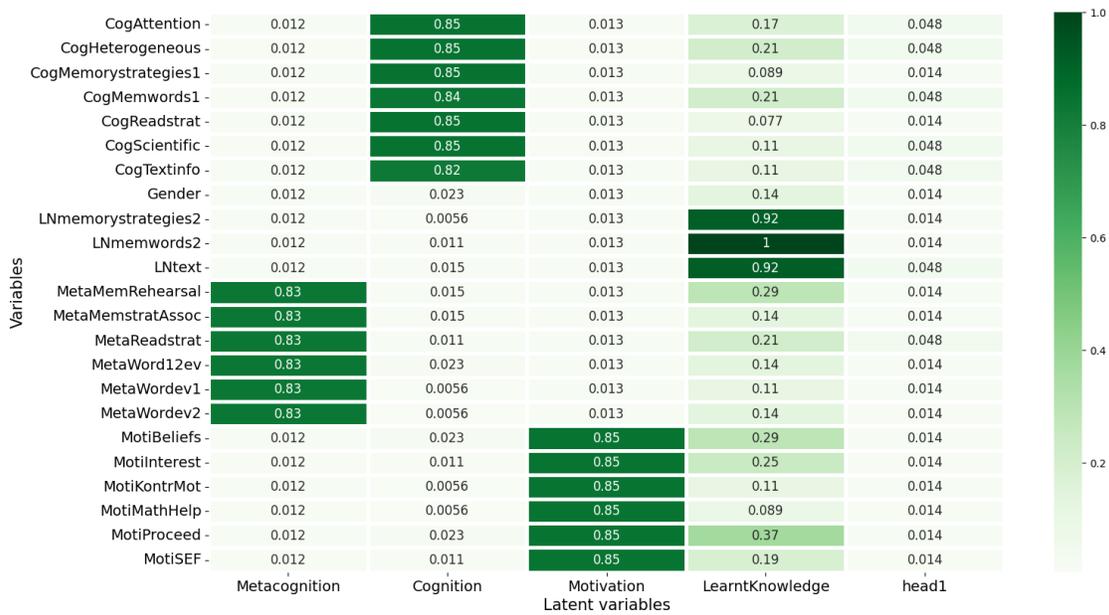

(b)

**Fig. 9** Knowledge extraction from the NSAI model for predicting factual knowledge (MF) using the balanced dataset: a) high-level visualization, and b) low-level visualization

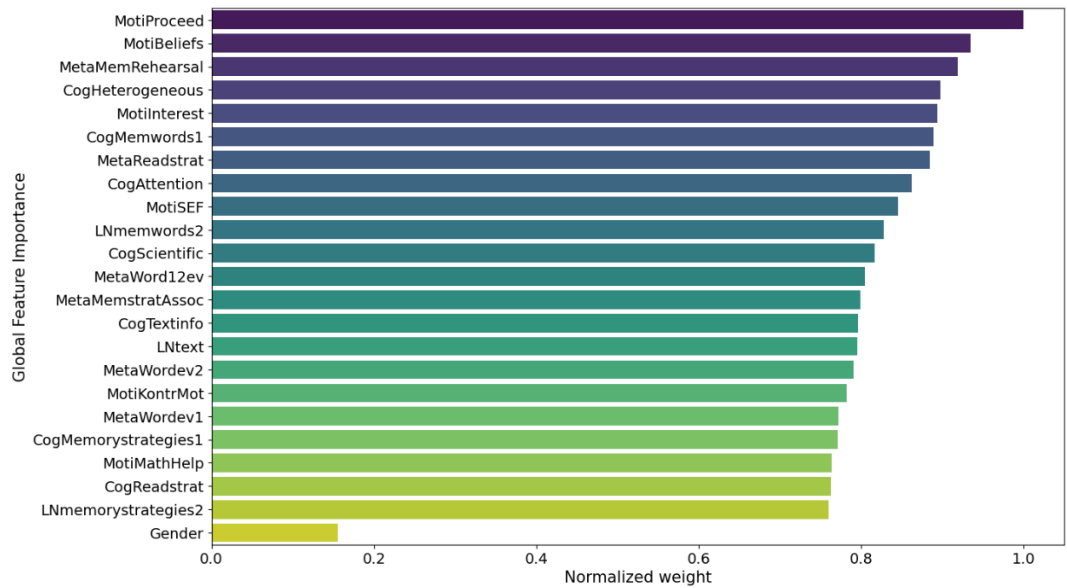

**Fig. 10** Global explanation of the learned NSAI model for predicting factual knowledge (MF) using the balanced dataset

### 4.3. Experiment two: Imbalanced data

This experiment evaluates the performance of symbolic, sub-symbolic, and neural-symbolic AI models under imbalanced data conditions, where low performers are underrepresented compared to high performers. Such an imbalance presents challenges for models, as they may favour high performers due to their larger presence in the dataset, potentially leading to lower recall for low performers. Below, we analyse how each AI family performs across factual and conceptual knowledge classification.

### 4.3.1. Quantitative comparison

Table 5 presents the performance of various models on the imbalanced dataset across different evaluation metrics. Since high performers make up the majority class, most models tend to favour them, resulting in low recall for struggling students. Overall, random forest achieves the highest AUC among the symbolic methods, while neural networks lead among the sub-symbolic methods. While both models show very good recall for high performers, they struggle with recall for low performers under standard classification thresholds. On the other hand, the NSAI model achieves the highest AUC scores across both knowledge types, demonstrating its discriminatory power in an imbalanced dataset. With competitive accuracy and the best recall, NSAI effectively distinguishes between low and high performers while maintaining a balanced trade-off between precision and recall.

**Table 5** Performance of various AI methods in predicting both factual and conceptual knowledge in imbalanced unseen test dataset

| AI family | methods | Metrics | Factual knowledge (MF) | | Conceptual knowledge (MC) | |
|---|---|---|---|---|---|---|
| | | | Low Performer | High Performer | Low Performer | High Performer |
| Symbolic | Decision tree | AUC (%) | 73 | | 68 | |
| | | Accuracy (%) | 78 | | 74 | |
| | | Precision (%) | 67 | 82 | 67 | 74 |
| | | Recall (%) | 48 | 89 | 8 | 98 |
| | Rule induction | AUC (%) | 69 | | 54 | |
| | | Accuracy (%) | 76 | | 74 | |
| | | Precision (%) | 62 | 78 | 54 | 76 |
| | | Recall (%) | 32 | 92 | 24 | 92 |
| | Random forest | AUC (%) | 78 | | 68 | |
| | | Accuracy (%) | 76 | | 74 | |
| | | Precision (%) | 80 | 76 | 57 | 75 |
| | | Recall (%) | 16 | 98 | 16 | 95 |
| Sub-symbolic | Logistic regression | AUC (%) | 75 | | 65 | |
| | | Accuracy (%) | 74 | | 69 | |
| | | Precision (%) | 67 | 82 | 54 | 84 |
| | | Recall (%) | 32 | 82 | 48 | 71 |
| | Bayesian network | AUC (%) | 75 | | 62 | |
| | | Accuracy (%) | 78 | | 72 | |
| | | Precision (%) | 69 | 79 | 50 | 74 |
| | | Recall (%) | 36 | 93 | 12 | 95 |
| | Neural network | AUC (%) | 76 | | 68 | |
| | | Accuracy (%) | 76 | | 69 | |
| | | Precision (%) | 62 | 78 | 36 | 74 |
| | | Recall (%) | 32 | 92 | 16 | 89 |
| Neural-symbolic | NSAI | AUC (%) | 79 | | 71 | |
| | | Accuracy (%) | 84 | | 68 | |

|  |  | Precision (%) | 71 | 88 | 45 | 88 |
|  |  | Recall (%) | 68 | 89 | 76 | 65 |

*Symbolic models.* For *factual knowledge*, symbolic models demonstrate strong accuracy, with decision trees achieving 78% and rule induction 76%. However, this performance is largely driven by their effectiveness in classifying high performers while struggling students remain difficult to detect. Decision trees excel for high performers (precision 82%, recall 89%) but capture only 48% of low performers, missing more than half of struggling students. Rule induction performs even worse, with a recall of just 32% for low performers, making it ineffective for early intervention. Random forests improve precision for low performers (80%), but their recall remains critically low (16%), meaning they miss most struggling students despite their high precision. AUC, which provides a more reliable assessment in imbalanced datasets by evaluating the model's overall discriminatory power, further highlights random forests as the strongest symbolic model, achieving the highest AUC (78%) and demonstrating superior classification capability. For *conceptual knowledge*, decision trees maintain high accuracy (74%) but struggle significantly with recall for low performers (8%), failing to provide meaningful early detection. Rule induction slightly improves recall (24%) but remains largely ineffective, as reflected in its low AUC (54%). Random forests, like decision trees, achieve similar accuracy (74%), indicating that while they are good at predicting high performers, they fall short in detecting low performers. However, their AUC (68%) reveals a better overall ability to distinguish between performance levels compared to other symbolic models.

*Sub-symbolic models.* For *factual knowledge*, sub-symbolic models demonstrate strong accuracy, with Bayesian networks achieving the highest (78%), followed by neural networks and logistic regression. However, this accuracy is largely driven by their ability to correctly classify high performers while misclassifying low performers. Both logistic regression and Bayesian networks struggle to identify struggling students, capturing only 32% and 36%, respectively, making them ineffective for early intervention. While all sub-symbolic models excel at identifying high performers (with more than 80% recall), their low-performer recall remains a critical weakness. AUC, which is the most important metric for imbalanced datasets as it measures the model's ability to discriminate between classes across different thresholds, provides a clearer picture of these models' classification performance. Neural networks achieve the highest AUC (76%), followed closely by Bayesian networks and logistic regression (both at 75%). These scores indicate that, despite their limitations in recall for struggling students at standard classification thresholds, sub-symbolic models still have strong overall discriminatory power. In *conceptual knowledge*, sub-symbolic models demonstrate moderate accuracy, with Bayesian networks achieving 72%, while logistic regression and neural networks reach 69%. However, this performance is largely driven by their ability to classify high performers correctly, while struggling students remain difficult to detect. Logistic regression improves recall for low performers (48%), but its overall classification ability remains limited. Bayesian networks and neural networks perform poorly in identifying struggling students, capturing only 12% and 16%, respectively, reinforcing their bias towards high performers. On the other hand, AUC provides a clearer assessment of these models' performance. Neural networks achieve the highest AUC (68%), compared to Bayesian networks and logistic regression, indicating better differentiation between student performance levels despite their poor recall at standard classification thresholds.

*Neural-symbolic model.* Among all models, NSAI demonstrates the strongest overall performance for both factual and conceptual knowledge. For *factual knowledge*, it achieves the highest accuracy of 84%, effectively classifying both high and low performers while maintaining a strong balance between precision and recall. In *conceptual knowledge*, it continues to outperform other models in detecting struggling students, ensuring better identification of at-risk learners compared to symbolic and sub-symbolic approaches. AUC underscores NSAI's advantage, achieving the highest scores for both factual (79%) and conceptual knowledge (71%), confirming its superior ability to distinguish between high and low performers while maintaining a balanced trade-off between recall and precision.

### 4.3.2. Qualitative comparison

The quantitative analysis in the previous section demonstrated that, except for the NSAI model, none of the other methods achieved at least 50% recall for low performers in either factual or conceptual knowledge under the standard classification threshold. Since high performers make up the majority class, most models tend to favour them, leading to high precision and recall for high performers but consistently low recall for struggling students. AUC analysis further reinforced NSAI's advantage, showing that regardless of classification thresholds, it outperformed all other methods in distinguishing between high and low performers. This indicates that NSAI not only maintains a better balance between precision and recall but also demonstrates stronger generalizability across different performance levels. Given these findings, models with weak predictive power may generate misleading patterns, as they primarily reflect the learning attributes of high performers while failing to capture crucial characteristics of struggling students. To ensure meaningful insights, this section qualitatively interprets only the NSAI model, as its predictions and learned patterns offer the most reliable foundation for identifying diverse learning profiles and supporting both struggling and high-performing students.

Fig. 11a and 11b illustrate the high-level knowledge extracted from the NSAI model for predicting factual and conceptual knowledge in the imbalanced dataset. For factual knowledge (Fig. 11a), cognition is the most influential factor, followed by learnt knowledge, then motivation, with head1 and metacognition contributing the least. For conceptual knowledge (Fig. 11b), learnt knowledge is the most influential factor, followed by motivation and cognition, with metacognition and head1 having relatively lower influence.

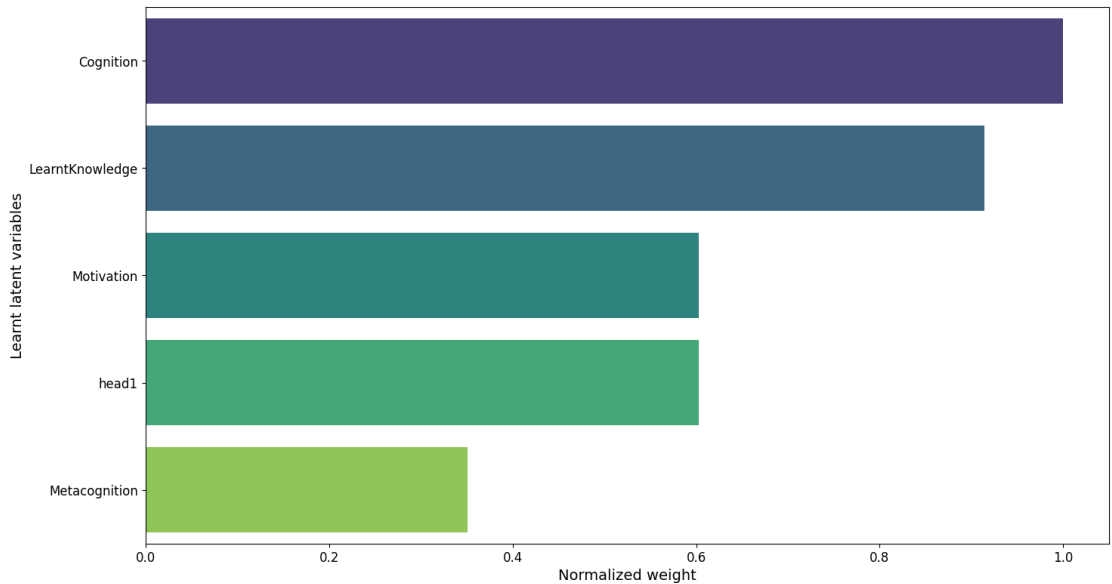

(a)

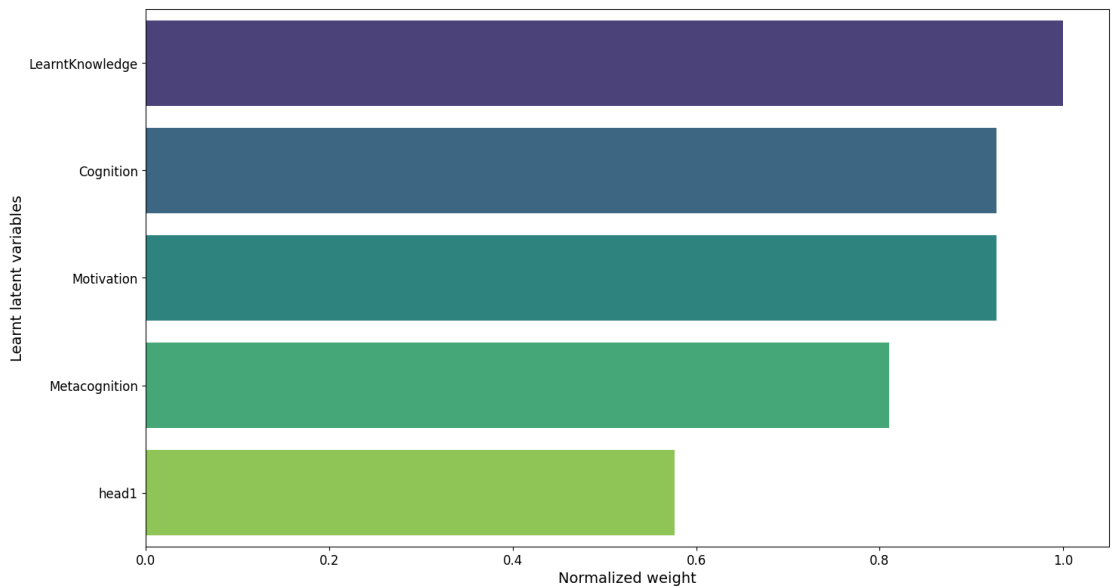

(b)

**Fig. 11** High-level knowledge extraction from the NSAI model for predicting: a) factual knowledge (MF), and b) conceptual knowledge (MC) using the imbalanced dataset

A more detailed visualization of the model (see Fig. 12a and 12b for factual and conceptual knowledge) reveals how different observed variables contribute to latent constructs in the imbalanced dataset. For factual knowledge, cognitive, motivational, metacognitive, and learnt knowledge-related attributes contribute to their respective constructs, highlighting their importance in factual knowledge acquisition. Scientific thinking level, identifying important information, attention, and heterogeneous thinking level show the highest contributions to head1, indicating that these latent variables capture aspects of *thinking level and information processing*. Fig. 13a confirms the contribution of these predictors by ranking them in the top five.

For conceptual knowledge, cognitive attributes similar to those in factual knowledge (in line with the injected rules) strongly load onto cognition, reinforcing their role in attention regulation, scientific reasoning, and memory-based learning strategies. Similarly, motivational, metacognitive, and learnt knowledge-related attributes contribute to their respective constructs, emphasizing their importance in conceptual knowledge acquisition. Interestingly, among the motivational factors, growth mindset plays the most significant role, followed by controlled motivation, proceeding with calculation, and seeking help, whereas autonomous motivation and self-efficacy ranked lower in the motivation construct. Three metacognitive factors—self-assessment, self-confidence in learning words I, and valuing association strategies—as well as cognitive association strategy use I, significantly contribute to motivation, suggesting a reinforcing relationship between metacognitive and cognitive strategy use and motivation in learning.

In terms of learnt knowledge, the two highest-ranked factors were text comprehension and memorized words II, followed by association strategy use II. Beyond gender and cognitive factors such as attention and heterogeneous thinking level, motivational and metacognitive aspects—particularly growth mindset and self-confidence in learning words II—also influenced learnt knowledge, underscoring their relevance in conceptual knowledge retention. Gender, attention, heterogeneous thinking level, and scientific thinking level showed the highest contributions to head1, suggesting that these latent variables capture key aspects of thinking level, information processing, and individual differences in conceptual processing. According to the global feature importance analysis (Fig. 13b), the top five predictors of conceptual knowledge were scientific thinking level, attention, heterogeneous thinking level, growth mindset, and self-evaluation. Additionally, among metacognitive strategies, self-evaluation, valuing association strategies, self-confidence in learning words I and II ranked the highest, while valuing rehearsal strategies contributed the least to metacognition. These highlight a few ways in which NSAI effectively refines the weights for each construct factor, enabling further theoretical refinement and optimization of the initially injected domain knowledge.

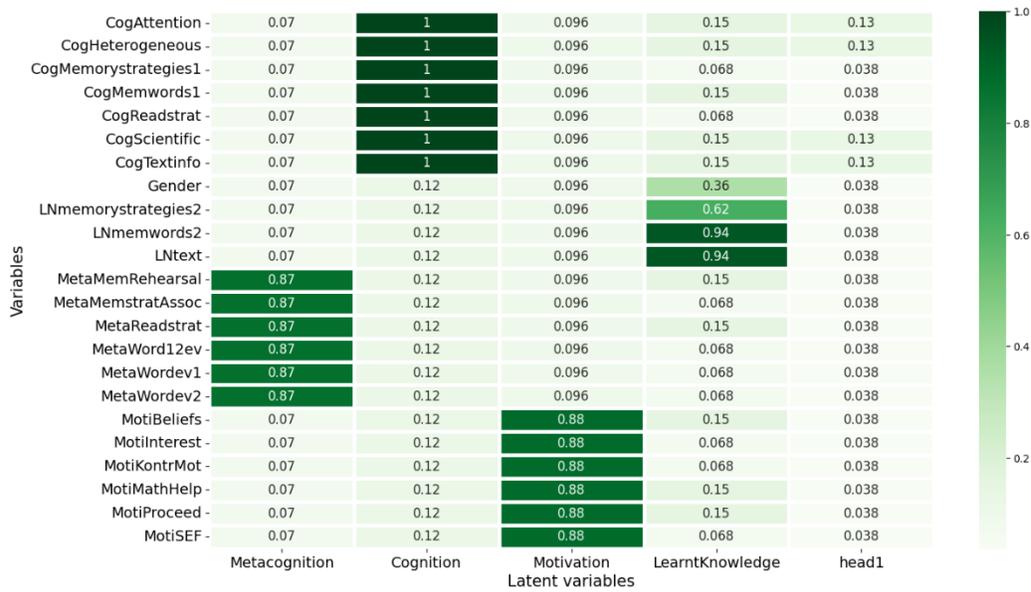

(a)

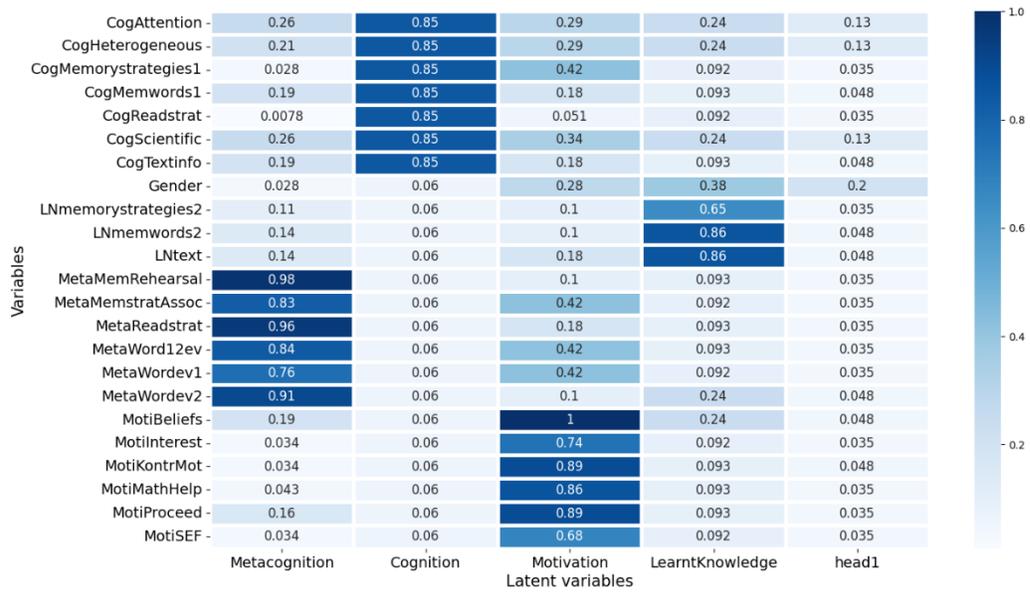

(b)

**Fig. 12** Low-level knowledge extraction from the NSAI model for predicting: a) factual knowledge (MF), and b) conceptual knowledge (MC) using the imbalanced dataset

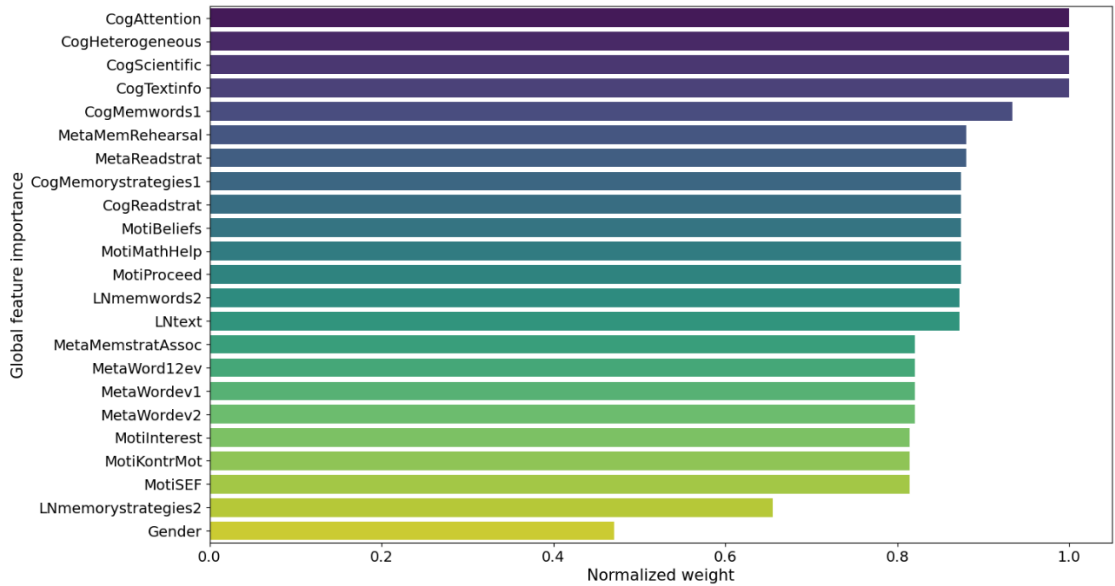

(a)

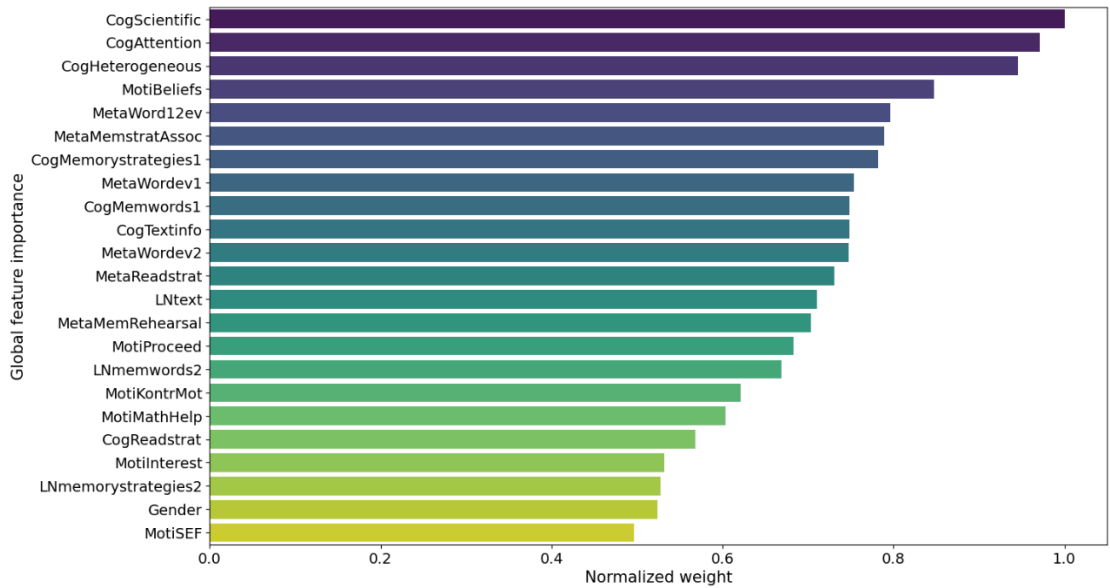

(b)

**Fig. 13** Global explanation of the learned NSAI model for predicting: a) factual knowledge (MF), and b) conceptual knowledge (MC) using the imbalanced dataset

## 5. Discussion and Conclusions

Given the increasing demand for responsible and trustworthy AI in education, this study evaluates symbolic, sub-symbolic, and neural-symbolic AI in educational data mining, considering both generalizability and interpretability. Using SRL indicators and learnt knowledge from Grade 3, it compares predictions of Grade 7 national math test outcomes, demonstrating how neural-symbolic AI balances these aspects while aligning with educational standards.

Our findings from the experiments on generalizability (or predictive performance) on the balanced dataset reveal that NSAI achieves one of the most well-balanced performances, making it amongst the most effective for identifying patterns relevant to educational interventions. On the imbalanced dataset, most symbolic and sub-symbolic models appeared to have favoured high-performing students, leading to a low recall for struggling learners. In contrast, NSAI effectively balances recall with competitive precision on both standard and various classification thresholds, making it the most reliable model for predicting both factual and conceptual mathematical knowledge in an imbalanced setting. The disparity in model performance between balanced and imbalanced datasets arises because most models, especially symbolic and sub-symbolic ones, tend to optimize for the majority class, leading to high precision but poor recall for struggling students in the imbalanced setting. In contrast, NSAI effectively mitigates this issue by leveraging knowledge injection and latent variable modelling, allowing it to generalize better across both groups and maintain high recall without sacrificing precision. Unlike the study conducted by Wnek and Michalski (1994), which showed that symbolic methods such as decision trees and rule-based learning outperformed neural networks in learning disjunctive normal form-type concepts from both noisy and noise-free data, our findings indicate that some sub-symbolic methods performed better than symbolic ones, particularly on balanced datasets. This aligns with studies like Hooshyar et al. (2019) and Ibrahim and Rusli (2007), which found that deep neural networks outperformed other methods in student performance prediction. Additionally, our findings align with Barbiero et al. (2023), Hooshyar et al. (2024), and Tato and Nkambou (2022) indicating that NSAI methods demonstrate better generalizability than other approaches, including deep neural networks. Similarly, our research shows that NSAI outperforms all symbolic and sub-symbolic models in predictive power.

With respect to interpretability, our results show that NSAI provides a more holistic and reliable framework for understanding student learning patterns compared to symbolic and sub-symbolic models. On a balanced dataset, symbolic models primarily emphasize cognitive factors such as attention and working memory skills, along with some motivational factors like controlled motivation and growth mindset. In contrast, sub-symbolic models, particularly logistic regression and neural networks, heavily rely on gender and heterogeneous thinking levels, often overlooking metacognitive aspects. In contrast, in both balanced and imbalanced datasets, NSAI effectively captures interactions between cognitive, metacognitive, motivational, and learnt knowledge factors, providing a deeper understanding of student learning patterns.

For factual knowledge in the balanced dataset, the NSAI method indicates that learnt knowledge, metacognition, and motivation contribute the most, followed by cognition with the least contribution among the constructs. However, in the global explanation, motivation and metacognitive factors emerge as the strongest predictors, with multiple predictors ranking among the top factors in factual knowledge acquisition. In the imbalanced dataset, cognition shows the highest contribution, followed by learnt knowledge, while metacognition and motivation contribute the least. Overall, based on global feature importance, cognition and learnt knowledge are the most influential, followed by motivation. In contrast, for conceptual knowledge, learnt knowledge surpasses cognition, and motivation becomes more significant. A closer look at NSAI's extracted rules highlights that among motivational factors, a growth mindset plays a central role, while in metacognition, valuing effective reading strategies, valuing rehearsal strategies, and self-evaluation are the most significant. Additionally, memorized words II and text comprehension emerge as the strongest predictors of learnt knowledge, reinforcing their role in knowledge retention. Unlike symbolic and sub-symbolic models, NSAI optimizes factor weights, ensuring better theoretical alignment and refinement of injected knowledge (Ciatto et al., 2024; Hooshyar & Yang, 2021, 2024). One reason for the discrepancy between the patterns identified and used in models to achieve predictions could be the limitations of post-hoc explanation methods like SHAP, which may not fully capture the actual decision-making process of complex models like neural networks (d'Avila Garcez & Lamb, 2020; Hooshyar & Yang, 2024; Lakkaraju & Bastani, 2020). For instance, neural networks and logistic regression—when explained using these methods—highlighted gender as the most important factor, contradicting other models that emphasized cognitive, metacognitive, and motivational aspects. A second reason could be the fundamental differences in how symbolic, sub-symbolic, and neural-symbolic models learn and represent information (Platzer, 2024). Symbolic models rely on explicit rules and structured reasoning, while sub-symbolic models, like deep neural networks, rely on statistical correlations and pattern recognition, making them prone to picking up dominant features that may not have causal relevance (Hooshyar et al., 2024). Neural-symbolic models, by integrating structured domain knowledge with data-driven learning, mitigate this issue by aligning extracted patterns with theoretical constructs, ensuring that identified factors reflect meaningful relationships rather than spurious correlations (Tato & Nkambou, 2022). The strong performance of neural-symbolic models can be attributed to their ability to integrate domain-specific rules, which act as an implicit regularization mechanism (Kärkkäinen, 2002). Overall, this structured integration results in a more constrained and interpretable model compared to the other two experimented families.

As our study reveals, overall, the NSAI method discovers semantically meaningful logic rules that go beyond the feature importance explanations provided by post-hoc methods like SHAP, which can only partially explain black-box models (d'Avila Garcez & Lamb, 2020; Hooshyar & Yang, 2024; Kindermans et al., 2019). Unlike symbolic or sub-symbolic models, NSAI offers a theoretical framework detailing how observable variables influence predictions while also explaining patterns learned through latent variables and nodes—an area where traditional symbolic and sub-symbolic approaches struggle. Notably, as the extracted knowledge aligns with the injected theory-based knowledge, NSAI's explanations clearly highlight key areas for refining initial educational knowledge. For instance, while we previously understood that metacognitive and motivation-related variables were relevant within each construct, we have now gained deeper insights into which specific learning competence factors play a more crucial role in predicting mathematical factual and conceptual knowledge. Additionally, NSAI reveals how these constructs and their factors contribute to what students have learned during test completion (acquired knowledge). Beyond the injected theoretical knowledge, NSAI also identified an additional latent variable, which we have named "thinking level and information processing." This variable emerges as a significant contributor to students' math learning, with its own set of specific influencing factors. This discovery enhances our understanding of learning dynamics, offering a more comprehensive, interpretable, and data-driven perspective on student performance. This collectively advances educational data mining by making it more interpretable, theory-driven, and trustworthy, bridging the gap between AI-driven insights and meaningful educational interventions.

When conducting experiments to evaluate the performance of symbolic, sub-symbolic, and neural-symbolic AI models under imbalanced data conditions (where low performers are underrepresented), significant advances in SRL can be expected. These advances would primarily focus on improving the ability of AI systems to support diverse learners, particularly those who perform poorly, by adapting learning experiences to meet their needs better. Firstly, the models hold the potential to significantly bolster the robustness of learner diversity. Symbolic AI could pave the way for clear, interpretable pathways for both high and low performers. It could also devise strategies to offer balanced support, even for those underrepresented in the data. Sub-symbolic models might use techniques like re-sampling to generalize better and support low performers. Neural-symbolic AI can learn complex patterns in low-performing learners' behaviour and integrate them into decision-making, offering more personalized learning trajectories. This potential for robustness enhancement in learner diversity is a promising aspect of AI models in education.

Secondly, the learning system's adaptivity can be significantly enhanced. With imbalanced data, models such as neural-symbolic systems can dynamically adjust learning content based on real-time learner progress, identifying when a low performer needs additional support or when a high performer might be ready for more challenging content. This level of adaptability reassures that more personalized learning strategies (e.g., re-reading, note taking) for each learner are possible, where the system can suggest alternative resources or adjust the pace based on performance levels. Furthermore, these AI models can enhance the personalization of learning strategies. Systems can now tailor the experience based on individual needs, such as providing more feedback, adjusting the difficulty level for low performers, and offering more complex challenges to high performers. Symbolic AI can create rules that trigger specific interventions, such as offering hints or changing task structures based on the learner's engagement and performance. At the same time, neural-symbolic models could identify patterns in learner behaviour that predict when low performers may need help and adjust accordingly. The feedback and monitoring mechanisms would also improve. By utilizing multimodal data (e.g., log files, concurrent verbalizations, eye movements) and analysing performance patterns, AI systems can provide immediate, targeted feedback to learners, particularly those struggling with specific tasks or concepts. With symbolic and neural-symbolic models, the feedback would not only help guide learners towards the correct answers but also support metacognitive awareness, which is the understanding and control of one's own thought processes, by explaining why specific strategies or approaches are effective.

Moreover, addressing fairness and bias reduction would be an important advancement. With imbalanced data, AI systems risk reinforcing biases that favour high performers. By focusing on fairness, these models would ensure that low performers are not marginalized but instead receive appropriate, tailored interventions, such as additional practice problems or personalized feedback, that help them progress in the same way high performers do. Symbolic models could be designed to ensure that low performers are given enough exposure to appropriate learning opportunities and are not excluded from the learning process. Additionally, the ability to track long-term performance would be significantly enhanced. Neural-symbolic models, for example, could monitor changes in a learner's behaviour over time, helping detect early signs of disengagement or difficulty and proactively providing support. This would result in continuous, dynamic adaptation to each learner's needs, helping prevent learners from falling behind. Finally, new learning models designed explicitly for underrepresented learners could emerge from this type of research. Symbolic models could be created to develop specific strategies and rules for low-performing learners, while neural-symbolic models could offer highly personalized, data-driven feedback and learning paths. These systems would provide support and engagement for low performers, which was difficult to achieve with traditional methods.

In conclusion, conducting experiments on symbolic, sub-symbolic, and neural-symbolic AI models under imbalanced data conditions could lead to significant advancements in self-regulated learning systems (Azevedo & Wiedbusch, 2023). These advancements include increased adaptivity, personalization, and fairness, ultimately ensuring low-performing learners receive the support they need to succeed. This approach, with its potential to foster a more equitable learning environment, offers hope for a future where all learners, regardless of their performance level, have opportunities to improve and succeed. Overall, our findings highlight the importance of hybrid neural-symbolic models in fostering responsible AI that supports data-driven yet explainable decision-making. As research in neural-symbolic AI evolves, future studies should explore how deeper integration of symbolic constraints and sub-symbolic learning can enhance trust, fairness, and effectiveness in educational AI applications.

### 5.1. Limitations and future work

This study has some limitations that open avenues for future research. Further experiments on different datasets and tasks—including those involving serious games, simulations, and sequential or multimodal trace data—could offer deeper insights into NSAI's adaptability. Additionally, exploring different variants of NSAI that inject symbolic educational knowledge in various ways, such as through inputs or loss functions, could help assess their impact on performance and interpretability. Future work should also involve a broader comparison of AI methods from various families to better understand their strengths and limitations in educational contexts. Lastly, a qualitative evaluation of interpretability involving stakeholders like teachers would be valuable for assessing how effectively different methods support human understanding and decision-making.

**Acknowledgements**

This work was supported by the Estonian Research Council grant (PRG2215).

Spirtes, P., & Glymour, C. (1991). An algorithm for fast recovery of sparse causal graphs. *Social Science Computer Review*, *9*(1), 62–72.

Tato, A., & Nkambou, R. (2022). Infusing expert knowledge into a deep neural network using attention mechanism for personalized learning environments. *Frontiers in Artificial Intelligence*, *5*, 921476.

Telesko, R., Jüngling, S., & Gachnang, P. (2020). *Combining Symbolic and Sub-symbolic AI in the Context of Education and Learning.* AAAI Spring Symposium: Combining Machine Learning with Knowledge Engineering (1).

Torralba, A., & Efros, A. A. (2011). *Unbiased look at dataset bias*. 1521–1528. https://doi.org/10.1109/CVPR.2011.5995347

Towell, G. G., & Shavlik, J. W. (1994). Knowledge-based artificial neural networks. *Artificial Intelligence*, *70*(1–2), 119–165.

Vellido, A., Castro, F., Nebot, A., & Mugica, F. (2006). *Characterization of atypical virtual campus usage behavior through robust generative relevance analysis*. 183–188.

Venugopal, D., Rus, V., & Shakya, A. (2021). *Neuro-symbolic models: A scalable, explainable framework for strategy discovery from big edu-data*. Proceedings of the 2nd Learner Data Institute Workshop in Conjunction with The 14th International Educational Data Mining Conference. https://ceur-ws.org/Vol-3051/LDI_4.pdf

Vincent-Lancrin, S., & Van der Vlies, R. (2020). *Trustworthy artificial intelligence (AI) in education: Promises and challenges*. https://doi.org/10.1787/19939019

Watts, T. W., Duncan, G. J., Siegler, R. S., & Davis-Kean, P. E. (2014). What's past is prologue: Relations between early mathematics knowledge and high school achievement. *Educational Researcher*, *43*(7), 352–360.46

Werder, K., Ramesh, B., & Zhang, R. (2022). Establishing data provenance for responsible artificial intelligence systems. *ACM Transactions on Management Information Systems (TMIS)*, *13*(2), 1–23.

White, A., & d'Avila Garcez, A. (2020). Measurable counterfactual local explanations for any classifier. In *ECAI 2020* (pp. 2529–2535). IOS Press.

Winne, P., & Azevedo, R. (2022). Metacognition and self-regulated learning. *The Cambridge Handbook of the Learning Sciences*, *3*, 93–113.

Wnek, J., & Michalski, R. S. (1994). Comparing symbolic and subsymbolic learning: Three studies. *Machine Learning: A Multistrategy Approach*, *4*, 489–519.

Yağcı, M. (2022). Educational data mining: Prediction of students' academic performance using machine learning algorithms. *Smart Learning Environments*, *9*(1), 11.

Yang, Y., Hooshyar, D., Pedaste, M., Wang, M., Huang, Y.-M., & Lim, H. (2020). Predicting course achievement of university students based on their procrastination behaviour on Moodle. *Soft Computing*, *24*, 18777–18793.

Yang, Y.-Y., Chou, C.-N., & Chaudhuri, K. (2022). Understanding rare spurious correlations in neural networks. *arXiv Preprint arXiv:2202.05189*. https://doi.org/10.48550/arXiv.2202.05189

Zhidkikh, D., Heilala, V., Van Petegem, C., Dawyndt, P., Jarvinen, M., Viitanen, S., De Wever, B., Mesuere, B., Lappalainen, V., & Kettunen, L. (2024). Reproducing Predictive Learning Analytics in CS1: Toward Generalizable and Explainable Models for Enhancing Student Retention. *Journal of Learning Analytics*, *11*(1), 132–150.

Zhidkikh, D., Saarela, M., & Kärkkäinen, T. (2023). Measuring self-regulated learning in a junior high school mathematics classroom: Combining aptitude and event measures in digital learning materials. *Journal of Computer Assisted Learning*, *39*(6), 1834–1851.

# Appendix A

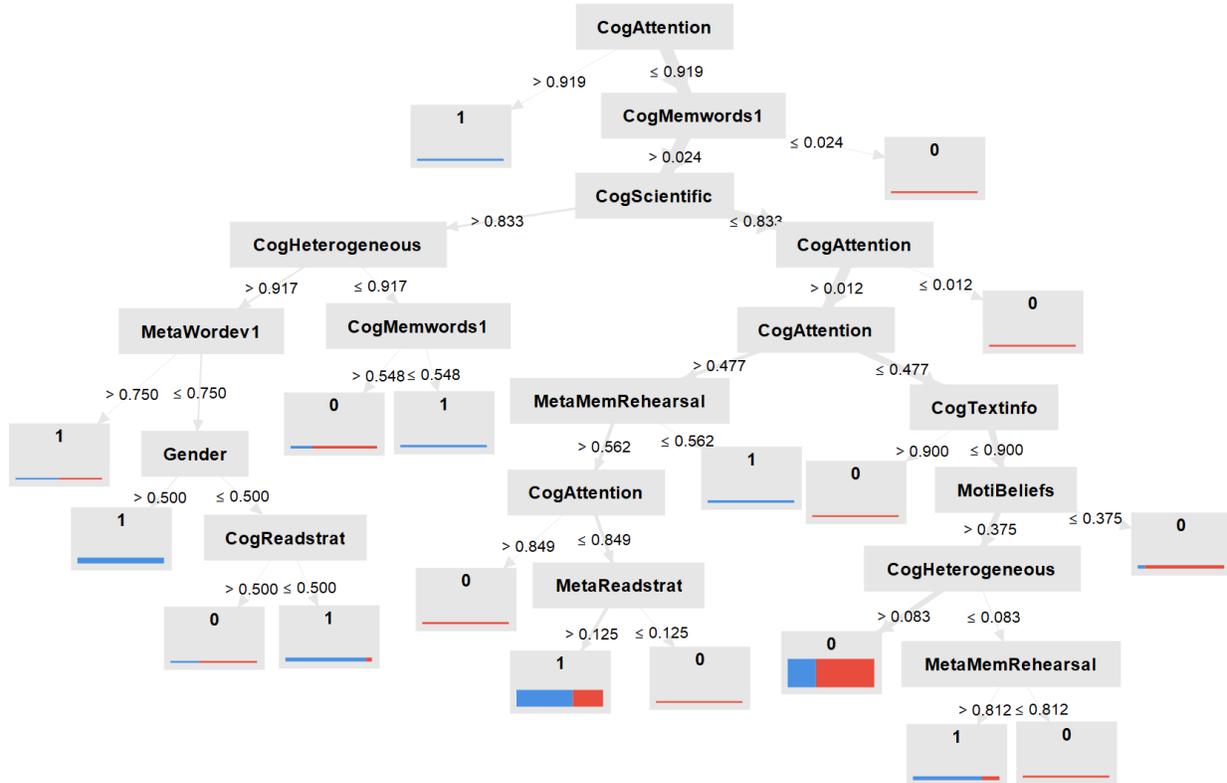

**Fig. 14** The decision tree learned for predicting conceptual knowledge (MC) using the balanced dataset

**Table 6** Rules learned for predicting conceptual knowledge (MC) using the balanced dataset

| |
|---|
| if CogAttention > 0.384 and CogScientific > 0.833 and CogHeterogeneous > 0.917 then 1 (33 / 4) |
| if CogAttention ≤ 0.360 and MetaReadstrat ≤ 0.438 and MotiInterest ≤ 0.812 then 0 (2 / 27) |
| if Gender ≤ 0.500 and CogAttention > 0.453 and MotiMathHelp ≤ 0.125 then 1 (18 / 1) |
| if LNtext ≤ 0.300 and Gender > 0.500 and MotiKontrMot ≤ 0.688 and CogHeterogeneous ≤ 0.750 and CogAttention ≤ 0.360 then 0 (1 / 23) |
| if MotiBeliefs > 0.625 and Gender ≤ 0.500 and MotiMathHelp > 0.375 and MotiInterest > 0.312 and MotiKontrMot > 0.354 then 1 (11 / 1) |
| if LNtext ≤ 0.300 and CogHeterogeneous ≤ 0.583 and MotiMathHelp ≤ 0.625 and MotiKontrMot ≤ 0.229 and CogHeterogeneous > 0.083 then 0 (3 / 20) |
| if MotiKontrMot ≤ 0.562 and CogHeterogeneous > 0.750 and Gender ≤ 0.500 and MotiMathHelp ≤ 0.708 then 1 (16 / 1) |
| if MotiKontrMot > 0.479 and LNmemwords2 > 0.405 and MotiMathHelp > 0.208 and MetaReadstrat ≤ 0.688 then 0 (0 / 19) |
| if CogAttention > 0.314 and MotiSEF ≤ 0.167 and CogAttention ≤ 0.500 then 1 (15 / 3) |
| if MotiMathHelp > 0.042 and CogMemwords1 > 0.690 and MetaMemstratAssoc ≤ 0.344 then 1 (15 / 2) |
| if CogAttention ≤ 0.477 and MotiMathHelp ≤ 0.708 and MotiMathHelp > 0.458 then 0 (2 / 14) |
| if CogMemorystrategies1 > 0.500 and CogAttention > 0.500 then 1 (12 / 1) |
| if MotiMathHelp ≤ 0.292 and CogAttention > 0.337 and LNmemorystrategies2 ≤ 0.500 and MetaMemstratAssoc > 0.344 then 0 (3 / 10) |
| if MotiMathHelp > 0.375 and CogAttention ≤ 0.442 and CogAttention > 0.198 then 1 (9 / 1) |
| if MotiBeliefs ≤ 0.625 and MotiMathHelp ≤ 0.083 then 0 (0 / 12) |
| if CogReadstrat ≤ 0.500 and MotiInterest > 0.562 and MotiKontrMot ≤ 0.146 then 1 (8 / 2) |
| if CogReadstrat > 0.500 and LNtext ≤ 0.300 then 0 (0 / 8) |
| if MotiKontrMot > 0.521 and MetaMemstratAssoc ≤ 0.594 then 1 (11 / 1) |
| if CogMemwords1 ≤ 0.690 and MotiKontrMot ≤ 0.354 and MetaMemstratAssoc ≤ 0.719 then 0 (1 / 13) |
| if MetaMemstratAssoc > 0.406 then 1 (21 / 7) |
| if CogHeterogeneous ≤ 0.583 then 0 (0 / 7) |
| if MotiInterest > 0.375 then 1 (3 / 0) |
| else 0  (0 / 0) |
| correct: 325 out of 361 training examples. |

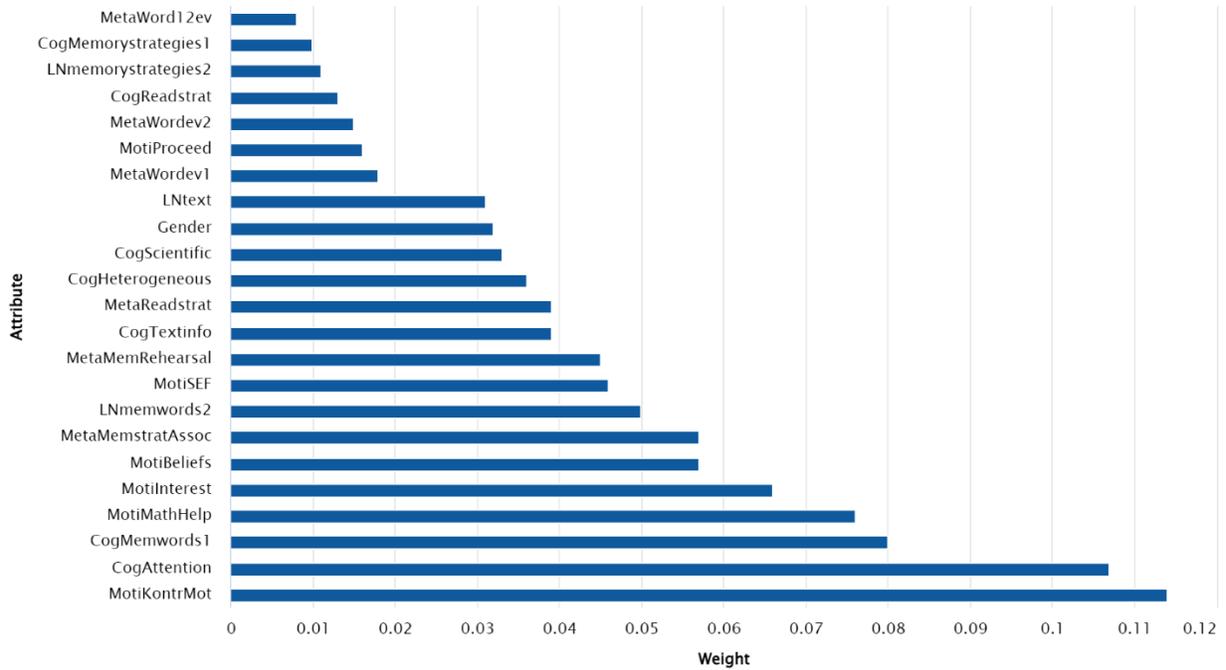

**Fig. 15** Feature importance of the learned random forest for predicting conceptual knowledge (MC) using the balanced dataset

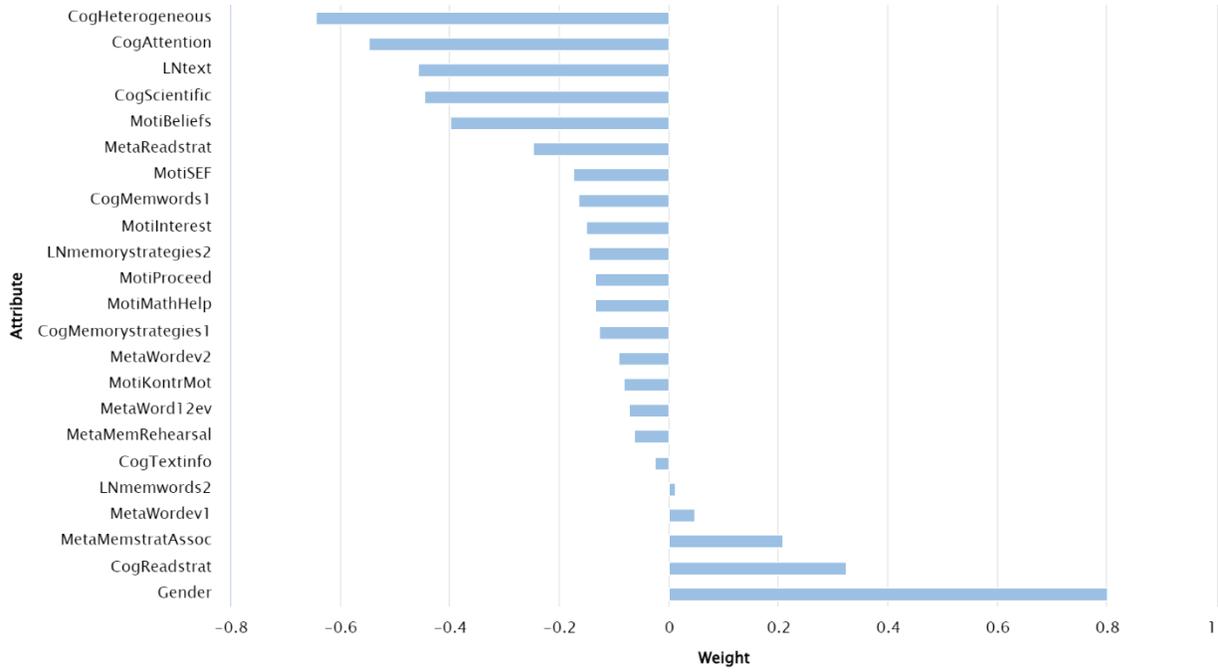

**Fig. 16** The attribute importance based on the learned logistic regression model learned for predicting conceptual knowledge (MC) using the balanced dataset

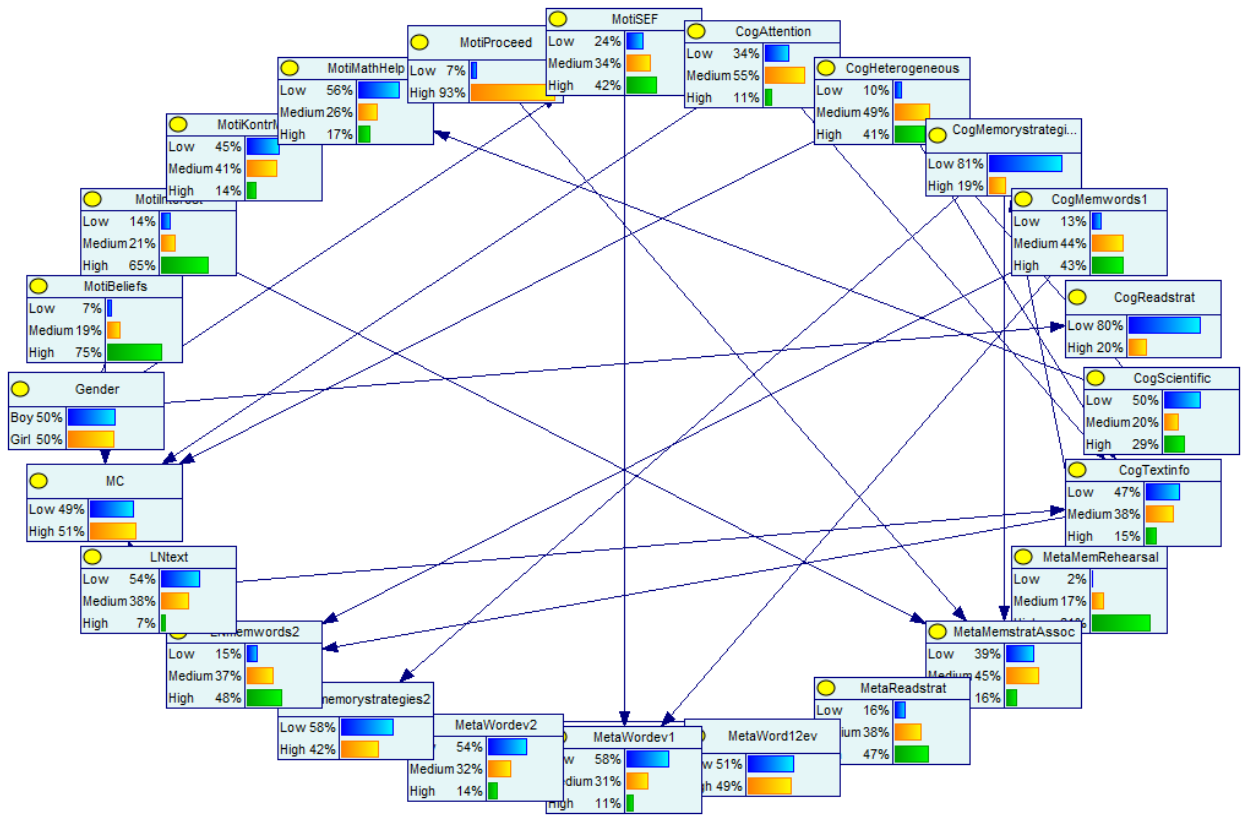

**Fig. 17** The Bayesian network learned for predicting conceptual knowledge (MC) using the balanced dataset

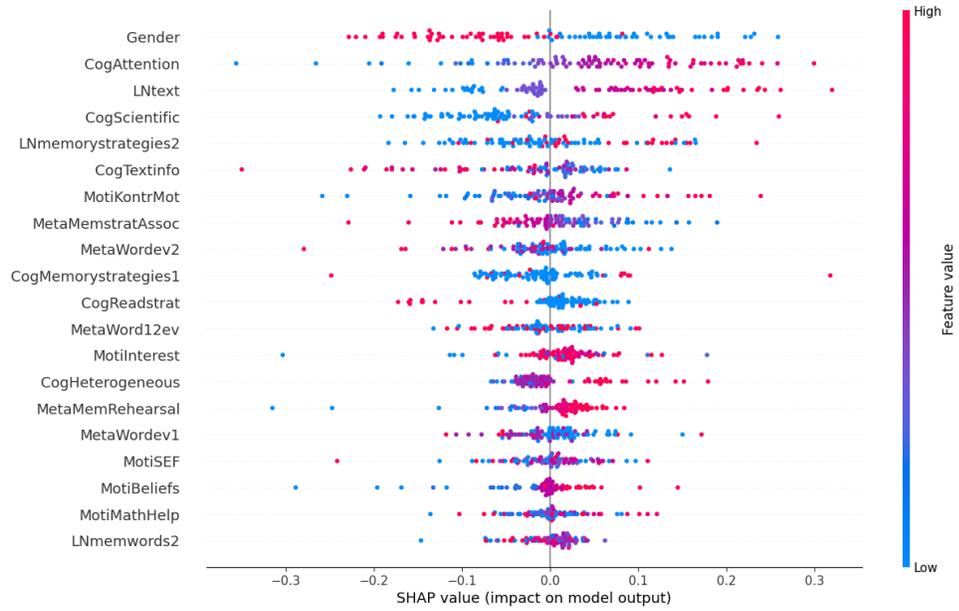

**Fig. 18** SHAP global explanation of the learned neural network for predicting conceptual knowledge (MC) using the balanced dataset.

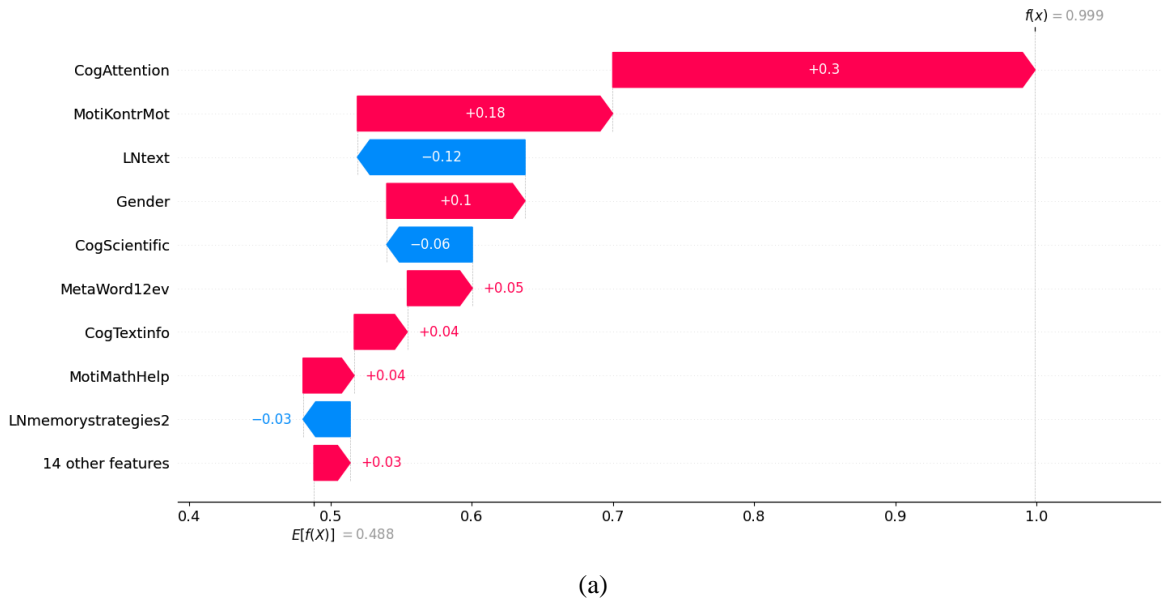

(a)

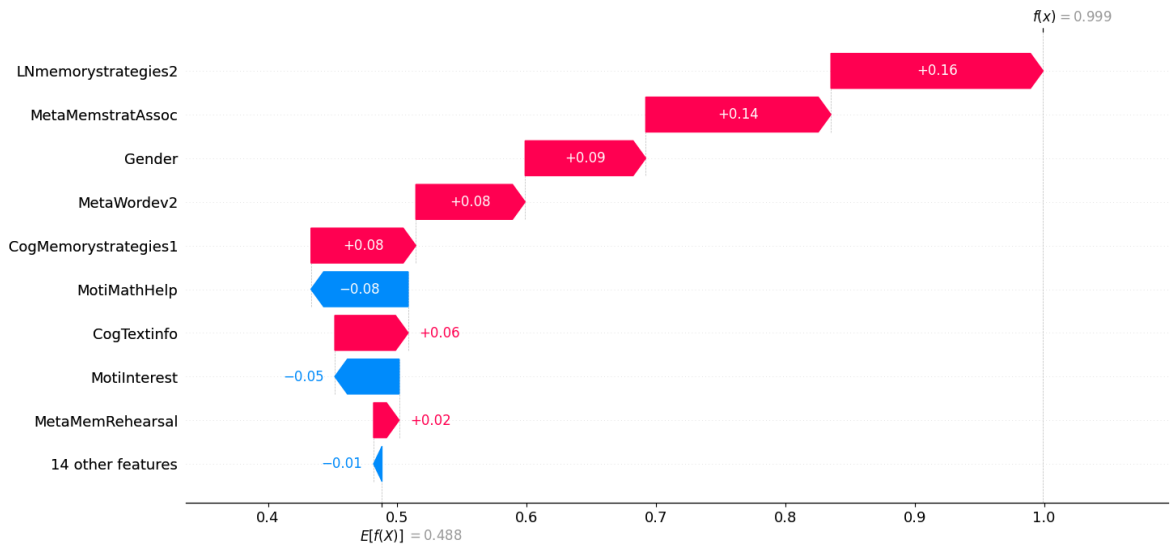

(b)

**Fig. 19** SHAP local explanation of the learned neural network for predicting conceptual knowledge (MC) using the balanced dataset: a) correct classified as high performer, and b) misclassified as high performer

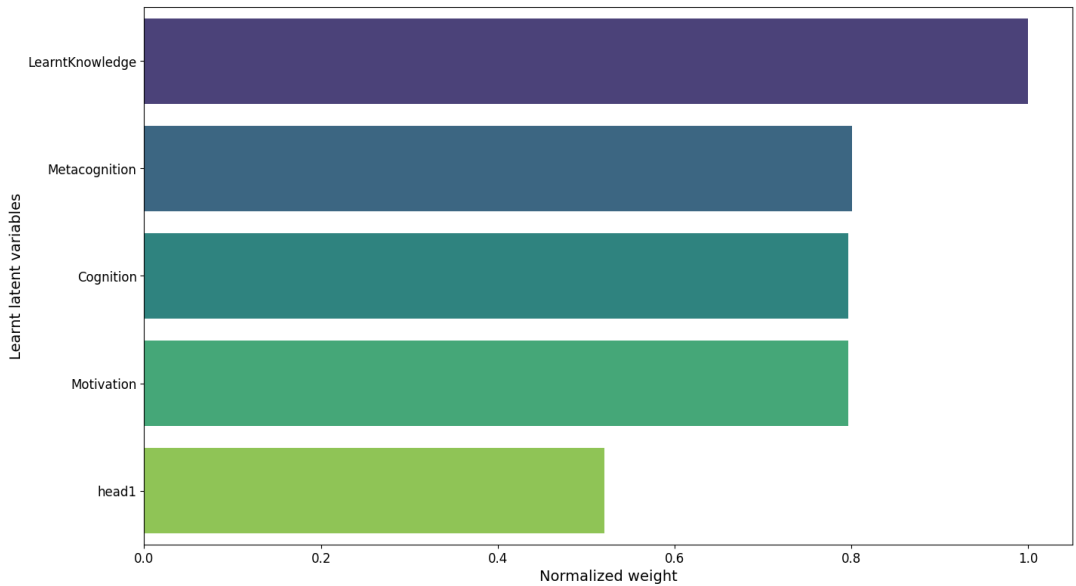

**Fig. 20** High-level knowledge extraction from the NSAI model for predicting conceptual knowledge (MC) using the balanced dataset

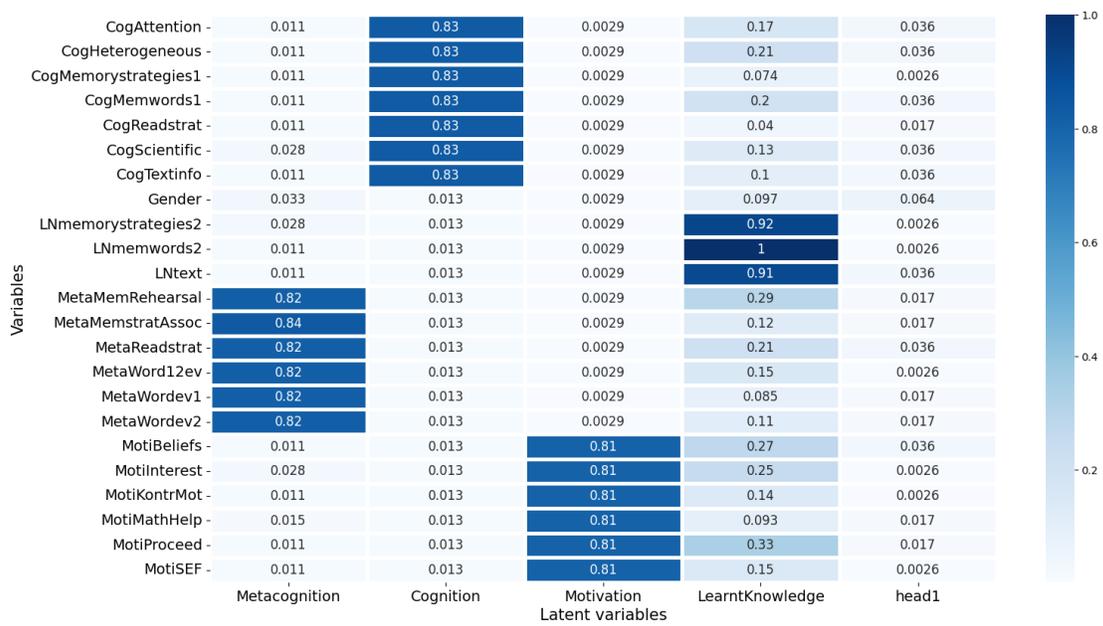

**Fig. 21** Low-level knowledge extraction from the NSAI model for predicting conceptual knowledge (MC) using the balanced dataset